%% file: main.tex
\documentclass[twoside,11pt]{article}
\usepackage{jair, rawfonts}
\usepackage{xcolor}

\usepackage{algorithm}

\usepackage{xspace}
\usepackage{theapa}
\usepackage{times}
\usepackage{multicol}
\usepackage{paralist}
\usepackage{tabularx}
\usepackage{listings}
\usepackage{amsmath,amssymb,amsfonts}
\usepackage{times}
\usepackage{graphicx}
\usepackage{subcaption}
\usepackage{float}
\usepackage{multirow}
\usepackage{boldline}
\usepackage{booktabs}
\usepackage{todonotes}
\usepackage{url}

\newcommand{\modelname}{{\sc ToolTango}\xspace}

\newcommand{\blue}[1]{{\leavevmode\color{black}{#1}}}

\jairheading{1}{1993}{1-15}{6/91}{9/91}
\ShortHeadings{Commonsense Generalization in Predicting Tool Interactions}
{Tuli, Bansal, Paul, \& Mausam}
\firstpageno{1}

\begin{document}

\title{\modelname: Common sense Generalization in Predicting Sequential Tool Interactions for Robot Plan Synthesis}

\author{\name Shreshth Tuli\thanks{Most work done when the authors were undergraduate students at Indian Institute of Technology Delhi, India.} \email s.tuli20@imperial.ac.uk \\
       \addr Department of Computing\\
       Imperial College London, UK
       \AND
        \name Rajas Bansal$^*$ \email rajasb@stanford.edu \\
       \addr Department of Computer Science\\
       Stanford University, USA
       \AND
       \name Rohan Paul \email rohan@cse.iitd.ac.in \\
        \addr Department of Computer Science and Engineering / Yardi School of Artificial Intelligence\\
       Indian Institute of Technology Delhi, India 
       \AND
       \name Mausam \email mausam@cse.iitd.ac.in \\
       \addr Department of Computer Science and Engineering / Yardi School of Artificial Intelligence\\
       Indian Institute of Technology Delhi, India
       }


\maketitle

\begin{abstract}
Robots assisting us in environments such as factories or homes must learn to make use of objects as tools to perform tasks, 
for instance using a tray to carry objects. 
We consider the problem of learning commonsense knowledge of when a tool may be useful 
and how its use may be composed with other tools to accomplish 
a high-level task instructed by a human. 
Specifically, we introduce a novel neural model, termed \modelname, that first predicts the next tool to be used, and then uses this information to predict the next action. We show that this joint model can inform learning of a fine-grained policy enabling the robot to use a particular tool in sequence and adds a significant value in making the model more accurate.
\modelname encodes the world state, comprising objects and 
symbolic relationships between them, using a graph neural network and is trained using demonstrations from human teachers instructing a virtual robot 
in a physics simulator.
The model learns to attend over the scene using knowledge of the goal 
and the action history, finally decoding the symbolic action to execute. 
Crucially, we address generalization to unseen environments 
where some known tools are missing, but alternative unseen tools are present. 
We show that by augmenting the representation of the environment with pre-trained embeddings 
derived from a knowledge-base, the model can generalize effectively to novel environments.
Experimental results show at least
48.8-58.1\% absolute improvement over the baselines in predicting successful symbolic plans for a simulated mobile manipulator in novel environments with unseen objects. 
This work takes a step in the direction of  enabling robots to rapidly synthesize robust plans for complex tasks, particularly in novel settings.

\end{abstract}

\input{intro}

\input{related_work}

\input{problem_formulation}

\input{technical_approach}

\input{data_collection}

\input{experiments}

\section{Conclusions}
\label{sec:conclusion}

This paper proposes \modelname, a novel neural architecture that learns a policy to attain intended goals 
as tool interaction sequences leveraging fusion of semantic and metric representations, 
goal-conditioned attention, knowledge-base corpora.
\modelname is trained using a data set of human instructed robot plans with simulated 
world states in home and factory like environments. 
It uses an independent model that predicts likelihood scores for each tool at each time-step.
The imitation learner demonstrates accurate commonsense generalization to environments 
with novel object instances using the learned knowledge of shared spatial and 
semantic characteristics. 
It also shows the ability to adapt to erroneous situations and stochasticity in action execution.
Finally, \modelname synthesizes a sequence of tool interactions with a 
high accuracy of goal-attainment.

\acks{Mausam is supported by an IBM SUR award, grants by Google, Bloomberg and 1MG, Jai Gupta chair fellowship, and a Visvesvaraya faculty award by Govt. of India. Rohan Paul acknowledges support from Pankaj Gupta Faculty Fellowship and DST's Technology Innovation Hub (TIH) for Cobotics. Shreshth Tuli is supported by the President's PhD Scholarship at Imperial College London. We thank the IIT Delhi HPC facility and Prof. Prem Kalra and Mr. Anil Sharma at the CSE VR Lab for compute resources. We thank Mr. Pulkit Sapra and Prof. P. V. M. Rao for assistance with CAD model creation. 
We thank \citeA{puig2018virtualhome} for sharing the implementation for baseline comparison. We are grateful to anonymous turkers and student volunteers for assisting in the data collection.}

\vskip 0.2in
\bibliography{sample}
\bibliographystyle{theapa}

\appendix
\input{appendix}

\end{document}

%% file: intro.tex
\section{Introduction}\label{sec:introduction}





\begin{figure*}[t]
    \centering
    \includegraphics[width=\linewidth]{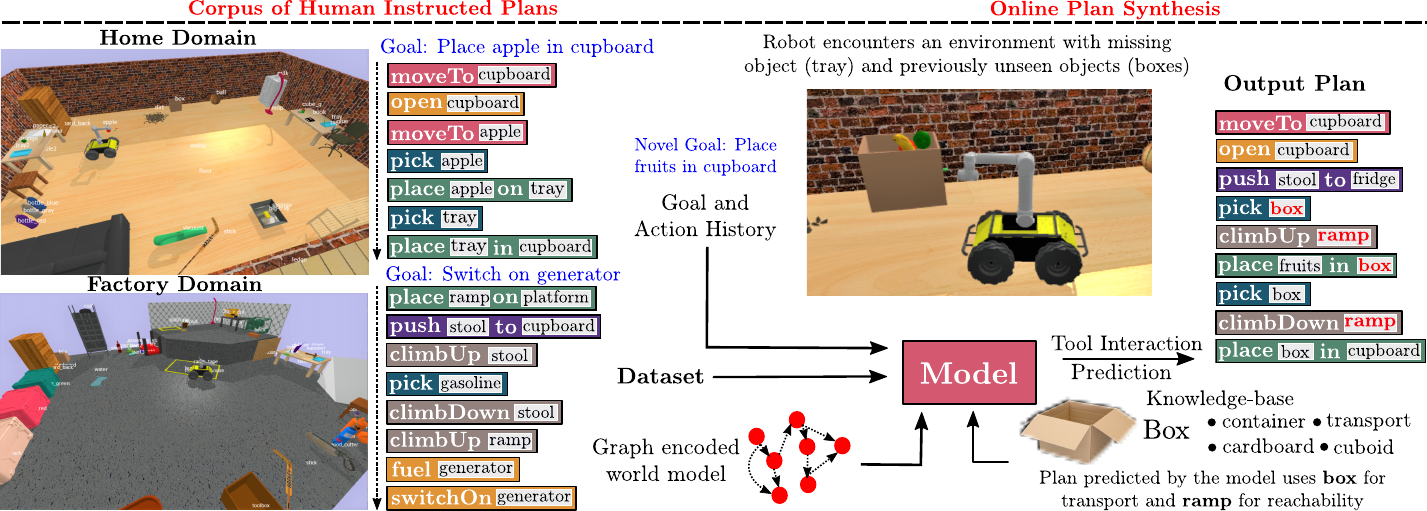}
    \caption{{
    \modelname acquires commonsense knowledge from human demonstrations leveraging graph-structured world representation, knowledge-based corpora and goal-conditioned attention to perform semantic tasks. Our aim is to acquire commonsense knowledge to develop a generalized goal-conditioned policy for a robot.
    }
    }
    \label{fig:motivation}
\end{figure*}

%
Advances in autonomy have enabled robots to enter human-centric domains
such as homes and factories where we envision them performing general purpose 
tasks such as transport, assembly, and clearing. 
Such tasks require a robot to interact with objects, 
often using them as \emph{tools}. 
For example, a robot asked to ``take fruits to the kitchen'', 
can use a \emph{tray} for carrying items, a \emph{stick} to fetch objects 
beyond physical reach and may use a \emph{ramp} to
reach elevated platforms.
Previous work has shown that the ability to predict the possible use of tools for a given 
task is often useful in guiding a robot's task planner 
towards plans likely to be feasible~\cite{driess2020deep,garrett2021integrated,choi2018creatingref2,levihn2014usingref3,fitzgerald2021modelingref8}.
In this work we consider the problem of predicting \emph{which} objects 
could be used as tools and \emph{how} their use can be 
composed for a task.  
In essence, we focus on the ability to predict appropriate tools that can guide the robot towards feasible and efficient plans and delegate the issue of dexterous tool manipulation to prior work~\blue{\cite{choi2018creatingref2,myers2015affordance}. }
Learning to predict task-directed tool interactions poses several challenges. 
First, real environments (a household or factory-like domain) 
are typically large where an expansive number of tool interactions may be possible 
(e.g., objects supporting others while transporting).  
%
Acquiring data for all feasible tool objects or exploring the space of tool interactions 
is challenging for any learner. 
Second, the usefulness of a tool varies with context.  
For example, placing milk in the cupboard may require the robot to elevate 
itself vertically using a ramp if the milk is placed 
at a height unreachable by the robot, but if the milk is kept on a table, 
a simple tray might suffice.
Third, the robot may encounter new environments populated with novel objects 
not encountered during training. 
Hence, the agent's model must be able to \emph{generalize} by reasoning 
about interactions with novel objects unseen during training.  
Humans possess innate commonsense knowledge about contextual use of tools for an intended goal~\cite{allen2019tools}. 
For example, a human actor when asked to move objects is likely to use trays, boxes, or even improvise with a new object with a flat surface. 
We aim at providing this commonsense to a robotic agent, so that it can generalize its knowledge to unseen tools, based on shared context and attributes of seen tools (see Figure~\ref{fig:motivation}). 

This paper takes a step in the direction of enabling robots to learn how to perform high-level tasks such as compositional tool use in semantic tasks, particularly in novel environments. 
This paper makes four main contributions.
As the first contribution, we present a crowd-sourced dataset of human-instructed plans where a 
human teacher guides a simulated mobile manipulator to leverage objects as tools to  perform multi-step actions such as assembly, transport 
and fetch tasks. 
The process results in a corpus of $\sim\!\!1,500$ human demonstrated robot plans 
involving diverse goal settings and environment scenes. 
%


Second, the dataset mentioned above is first used to  supervise a (1-step) neural imitation learner that predicts tool applicability given the knowledge of the world state and the intended goal.  We show how learning a dense embedding for the environment and that of a background knowledge base can enable the model to generalize to novel scenes with new object instances that may share semantic attributes with objects seen during training; a common problem in state of the art task and policy learning approaches.
We introduce a graph neural architecture, \textsc{ToolNet}, that   
encodes both the metric and relational attributes of the world state 
as well as available taxonomic resources such as $\mathrm{ConceptNet}$~\cite{speer2017conceptnet}. 
The \textsc{ToolNet} model predicts tool use by learning an attention over 
entities that can potentially serve as tools. 
Implicitly, the model acquires knowledge about primitive spatial characteristics 
(typically an output of a mapping system) and semantic attributes (typically 
contained in taxonomic resources) enabling generalization to novel contexts with previously unseen objects. 

Third, we present an imitation learner that uses the same dataset to make action predictions towards the intended goal.
We term the model, \underline{T}ool Inter\underline{a}ction Prediction \underline{N}etwork for \underline{G}eneralized \underline{O}bject environments (\textsc{Tango}). Similar to \textsc{ToolNet}, \textsc{Tango} also encodes the world state using a graph neural network and learns to attend over the scene using knowledge of the goal and the action history, 
finally decoding the symbolic action to execute. 
The action predictions are interleaved with physics simulation (or execution) steps, which ameliorates the need for modeling the complex effects of actions inherent in tool interactions.   

As a final contribution, we combine the two models \textsc{ToolNet} and \textsc{Tango} in a single architecture. Specifically, the joint model first makes the predictions for the next tool, and then uses this information to make a better action prediction. We show that this can inform learning of a fine-grained policy enabling the robot to use a particular tool in sequence and adds a significant value in making the model more accurate.
We term this joint model~\textsc{ToolTango}.

Experimental evaluation with a simulated mobile manipulator demonstrates (a) accurate prediction of tool interaction 
sequences with high executability/goal-attainment likelihood, (b) common sense generalization to novel scenes with unseen object instances, and (c) robustness to unexpected errors during execution.  Additionally, compared to \textsc{Tango}, we demonstrate the benefit of using \textsc{ToolNet} predictions for specific complex settings requiring multiple tools to reach the goal state.
Experiments show that in previously seen settings, \textsc{ToolTango} gives an absolute improvement of 3.38-5.59\% in reaching a goal state for a simulated mobile manipulator compared to the state-of-the-art \textsc{Tango} model. In unseen settings, the unified model gives 2.48-3.58\% improvement compared to \textsc{Tango}.
In comparison to a simple affordance prediction approach, the proposed model performs better in 
learning the pre-conditions for tool use. Further, the use of a neural model 
enables higher goal-reaching performance and faster training compared to a vanilla reinforcement learning approach.

A preliminary version of \textsc{ToolNet} that performs single tool prediction using only the initial environment state was presented as part of the Workshop on Advances \& Challenges in Imitation Learning in Robotics at Robotics Science and Systems (RSS) 2020 conference~\cite{toolnet}. Our \textsc{Tango} model was presented as part of the International Joint Conference in Artificial Intelligence (IJCAI) 2021~\cite{tango}.
This journal submission presents a substantially detailed exposition of the \textsc{ToolNet} and \textsc{Tango} models and includes additional supporting background material. Moreover, it also extends \textsc{ToolNet} to predict \emph{next} tool at each step of a multi-step action execution, instead of a single tool for the whole sequence, as in the original paper. This paper also combines the two models into \textsc{ToolTango}, and establishes its superior performance. 


The remainder of the paper is organized as follows. Section~\ref{sec:related_work} overviews related work. Section~\ref{sec:formulation} formulates the problem of predicting tool interaction sequences. Section~\ref{sec:model} details the technical approach and presents the \textsc{ToolTango} model. The data collection platform and the dataset in detailed in Section~\ref{sec:data}. Section~\ref{sec:experiments} provides the experimental details and results. Finally, Sections \ref{sec:limitations} and ~\ref{sec:conclusion} summarizes the work and lays 
out avenues for future work.
The associated code, data set and videos are available at \url{https://github.com/reail-iitd/tango}. Links to all supplementary material are given in Appendix~\ref{appendix:supp_material}.

%% file: related_work.tex
\section{Related Work}\label{sec:related_work}

\subsection{Classical Planning}

\blue{Reaching goal states from a given world state through symbolic actions is closely tied to the domain of  \textit{task and motion planning} (TAMP)~\cite{garrett2021integrated}. Literature presents a large volume of work in this domain, ranging from constraint satisfaction to search based methods. For instance, \citeA{toussaint2018differentiable} a planner to compose physics tool interactions 
using a logic-based symbolic planner. Similarly,~\citeA{srivastava2014combined} provide an implementation agnostic interface between a task and motion planner to combine both for goal-directed planning. \citeA{garrett2020pddlstream} extend PDDL descriptions to include generic and declarative specifications enabling the planning to be domain independent. These methods only utilize the task properties through action constraints. For instance, the knowledge that a tray can carry multiple objects is only encoded through the constraint that an object can be placed on it. This allows such methods to create \textit{any} feasible plan and not leverage commonsense knowledge to ensure that a goal state is reached in a few steps.  Some recent works such as by~\citeA{silver2021planning} aim to \textit{learn} object importance through human demonstrations. However, such methods can only work on objects seen previously in training and cannot generalize to unseen objects. We consider a part of the broad TAMP framework that focuses on determining the set of actions that are likely to take an agent to the goal state while taking the motion planning as a behavioural routine. We build our work on the observation that in domain with a large number of states and possible interactions, the task planning itself becomes challenging. We consider home and factory-like domains inspired from VirtualHome~\cite{puig2018virtualhome} and similar related works. We ensure that in our domains the number of objects is large and they can be contained within/supported or transported by other objects as tools (details in Section~\ref{sec:data_collection}). Similar to the work by~\citeA{silver2021planning}, our model learns to prune-away irrelevant objects, additionally considering a domain with richer inter-object interactions. Consequently, our learner makes additional use of semantic properties and exploits correlations between actions and outputs interactions that are likely to lead to successful plans.}

\subsection{Learning tool manipulation skills } 
Learning control policies for manipulating tools has received recent attention in 
robotics. 
\citeA{finn2017one} and \citeA{parkinferring} learn tool manipulation policies from  
human demonstrations. 
\citeA{xie2019improvisation} and \citeA{yildirim2015galileo} learn physics models and 
\emph{effects} enabling goal-directed compositional use. 
\citeA{liu2018physical} address the problem of learning primitive physical decomposition of tool like object through
its physical and geometric attributes enabling their human-like use. 
\citeA{wu2016physics} learn physical properties of objects from unlabeled videos. 
\citeA{nair2017combining} and \citeA{lynch2019learning} learn to interact with objects 
in a self-supervised setup. 
Efforts such as \citeA{holladay2019force}, and \citeA{antunes2015robotic} 
plan tool interactions modeling contact and force interactions. 
Another set of efforts focuses on incorporating the ability to discover the use of objects as tools and using them for plan completion. 
Rich symbolic architectures such as ICARUS~\cite{choi2018creatingref2}, KDAP~\cite{kroemer2012kernelref4} and ROAR~\cite{lee2015learningref11,levihn2014usingref3} attempt to model tool use via PDDL-like descriptions expressing 
the applicability and post effects of tool use. In particular, ICARUS~\cite{choi2018creatingref2} models bridge or staircase construction via lifted symbolic concepts which can be grounded to real or imagined objects in the environment. The framework presented in this paper takes inspiration from such classic works and builds on the use of learned representations for generalization to new scenes and objects. Further, instead of encoding such knowledge via a symbolic representation for each tool, the framework acquires such knowledge in a data-driven way from human demonstration.
Thus, the aforementioned works focus on learning \emph{how} to manipulate
a tool. Our paper considers the complementary problem of predicting 
\emph{which} objects may serve as tools for a given task 
while delegating the issue of tool manipulation to the 
works as mentioned earlier.

The works of~\citeA{fitzgerald2021modelingref8} and~\blue{\citeA{gajewski2019adaptingref9}} considers the specific problem of learning the physical motion of tools (such as spatial, hammer, mug, etc.) from kinesthetic demonstration trajectories. Further, the authors present a framework to learn corrections or replacements enabling improvisation when the actual task execution scenario differs from the taught demonstration. The focus of the work is in learning physical motion of tools for short range tasks. \blue{This work lifts the problem of dexterous tool manipulation and focuses on utilizing tools to reach goal states  (\textit{e.g.}, using a box to transfer multiple objects). This work also extends planning to leverage multiple tools to attain goals (using boxes and later using ramps). Rather than use fine-grained kinesthetic teaching, we instead use demonstrations of.long range plan executions.}  The two efforts are highly complementary. The tool-use trajectories learned by \blue{prior} work could potentially be used in our framework to accomplish long-range tasks. Similarly, such works can use the framework presented here to compose skills and generalize to unseen tools. 

\subsection{Learning symbolic action sequences}
Others address the problem of acquiring knowledge for 
completing high-level task specifications. 
\citeA{puig2018virtualhome,liao2019synthesizing} 
create a knowledge base of task decompositions as \emph{action sketches} and learn to translate 
sketches to executable plans. These efforts rely on the causal knowledge of sequences on 
sub-steps required to achieve an activity which are then contextually grounded. 
Instead, this work learns compositional tool use required to achieve the task without any
causal sequence as input.  
\citeA{huang2019neural} learn task decompositions from human demonstration videos. 
However, \blue{their} work does not explicitly model the physical constraints of the robot and does not 
generalize to new environments. 
\citeA{lee2015learningref11} and~\citeA{huang2015leveragingref12} learn trajectory-aware manipulation of deformable objects using a non-rigid registration method and human demonstrations. These methods can effectively handle visual variation in manipulating objects; however, they cannot be extended to generate action sequences to reach goal constraints given an unseen environment.
\citeA{shridhar2019alfred} take a similar approach by collecting natural language 
corpora describing high-level tasks and learn to associate instructions to 
spatial attention over the scene. 

Other works study \emph{relational} planning domains, where model is trained using RL on small problems, but tested zero-shot on large problems \cite{garg2019size,garg2020symbolic}. \citeA{sharma2022} extend this to imitation learning framework, but these works cannot perform tool  generalizations. 
\citeA{boteanu2015towards} present a symbolic system where a robot imitates demonstrations from a single teacher. In new environments, it adapts the plan by performing object replacements using ConceptNet relation edges. A rich set of approaches \textit{learn} affordances by mapping objects to preferred locations or learning the co-use of objects to accomplish a task. For instance,~\citeA{fitzgerald2018humanref10} provide a method of acquiring and transferring such knowledge to new tasks by learning to map between objects in the old and new environments in a task-dependent manner. This work does not explicitly attempt to learn such associations and presents a restricted generalization capacity. Instead, such learning is implicit in the learned policy that predicts a sequence of robot actions to achieve the goal while using tools (e.g., fetching the tool, using the tools and attaining its post effects).

Our approach draws inspiration from the above-mentioned works 
in that, we learn to predict tools that can be considered as  
sub-goals to guide planning for a high-level task. 
\blue{In comparison to these approaches, our method provides two key contributions.} First, we explicitly model the 
physical constraints arising from a mobile manipulator interacting in the work space.   
Second, instead of learning actions predicated on specific object instances, 
we address generalization to new object instances using 
primitive spatial and semantic characteristics. \blue{We propose} a neural model trained using a corpus of multiple and varied demonstrations provided by several teachers. Our model uses a dense embedding of semantic concepts, enabling generalization beyond relationships explicitly stored in ConceptNet.

\subsection{Commonsense knowledge in instruction following}
Acquisition of common sense knowledge has been previously explored for the task of 
robot instruction following~\cite{sarathy2018macgyverref5}.
\citeA{nyga2018grounding} present a symbolic knowledge base for procedural knowledge of tasks 
that is utilized for interpreting underspecified task instructions. 
Efforts such as \citeA{kho2014robo} propose a similar database encoding common sense knowledge 
about object affordances (objects and their common locations). 
Others such as \citeA{jain2015planit} learn motion preferences 
implicit in commands. 
\citeA{misra2016tell} ground instructions for recipe preparation tasks. 
Their model can generalize to new recipes, but only in environments with previously \emph{seen} objects. 
In contrast, our model generalizes to worlds with previously \emph{unseen} tools.  
Others, such as \citeA{nair2019autonomousref6} and \cite{nair2020featureref7} explore the problem of robot tool construction, \textit{i.e.}, creating tools from parts available in the environment. However, the limited set of commonsense concepts, such as attachment in~\cite{nair2019autonomousref6} restricts the space for generalization in such works. A rule-based approach typically does not scale well to develop a robust generalization model. We thus utilize a commonsense embedding based approach that utilizes ConceptNet vectors to encapsulate the concepts and generalize to unseen tools and object settings.

\citeA{chen2019enabling} present an instruction grounding model that leverages common sense taxonomic 
and affordance knowledge learned from linguistic co-associations. 
\citeA{Bisk2020} consider the problem of learning physical common sense
associated with objects and interactions required to achieve tasks
from language only data sets. They study this problem in the context
of question-answering to enable synthesis of textual responses that
capture such physical knowledge.
The aforementioned approaches predict latent constraints or affordances for a 
specified task. This work, additionally predicts the \emph{sequence} of 
tool interactions implicitly learning the causal relationships between tools use and effects. 
Specifically, this paper focuses on a learning common sense tool use in the context 
of following instructions that require multiple object interactions to 
attain the intended goal.

\subsection{Synthetic Interaction Datasets } 
Virtual environments have been used to collect human demonstrations for 
high-level tasks. 
\citeA{puig2018virtualhome} introduce a knowledge base of actions required to perform activities in a virtual home environment. 
\citeA{shridhar2019alfred} provide a vision-language dataset translating symbolic actions for a high-level activity to attention masks in ego-centric images. 
\citeA{nyga2018cloud} curated data sets that provide a sequence of \emph{How-To} 
instructions for tasks such as preparing recipes.
\citeA{myers2015affordance} present an affordance detection dataset for tool parts with geometric features.
Others such as \citeA{jain2015planit}, \citeA{scalise2018natural} and \citeA{mandlekar2018roboturk} present simulation environments and data sets for tasks such as learning spatial affordances, situated interaction or learning low-level motor skills. 
The present data sets possess two limitations that make them less usable 
for the learning task addressed in this work. 
First, the data sets are collected using human actors or avatars but 
do not explicitly model a robot in their environment. 
Though virtual agents 
serve as a proxy for the robot, they preclude modeling of the physical constraints and the range of tasks an robot can perform. 
Second, a majority of the data sets aim at visual navigation and limited physical interaction with objects. They are 
less amenable to interactions (e.g., containment, pushing and attachment etc.) inherent in tool use. 
Data sets utilized in robotic tool use literature, including the \blue{UMD Part Affordance Dataset}~\cite{myers2015affordance}, are confined mainly to local use, such as finding the appropriate tool part for tool use and manipulation. Such data sets are less amenable to multi-stage plans in large workspaces.

%% file: problem_formulation.tex
\section{Problem Formulation}\label{sec:formulation}
\subsection{Robot and Environment Model}
We consider a mobile manipulator operating in a known  
environment populated with objects. 
An object is associated with a pose, a geometric
model and symbolic states such as 
$\mathrm{Open/Closed}$, $\mathrm{On/Off}$ etc.  
We consider object relations such as (i) \emph{support} e.g., 
a block supported on a tray, (ii) \emph{containment}: items placed 
inside a box/carton and (iii) \emph{attachment}: a nail attached to a wall, 
and (iv) \emph{contact}: a robot grasping an object.   
Let $s$ denote the world state that maintains (i) metric information: 
object poses, and (ii) symbolic information:  object states, class type and 
object relations as $\mathrm{OnTop}$, $\mathrm{Near}$, $\mathrm{Inside}$ and 
$\mathrm{ConnectedTo}$. 
Let $s_{0}$ denote the initial world state and 
$\mathcal{O}(\cdot)$ denote a map from world state $s$ to 
the set of object instances $O = \mathcal{O}(s)$ populating state $s$. 
Let $\tau$ denote the set of tool objects that the robot can use in its plan. Note that only movable objects in the scene are 
considered as potential tools. Hence, $\tau  \subseteq  \mathcal{O}(s)$.  
Online, the robot may encounter \emph{unseen} objects in its environment. 

Let $A$  denote the robot's symbolic action space. An action $a\in A$ is abstracted as  $I(o^1, o^2)$, with
an action type predicate $I \in \mathcal{I}$ that affects the states of objects 
$o^1 \in O$ and $o^2 \in O$, for instance, $\mathrm{Move(fruit_{0}, tray_{0})}$. 
Here the arity of the interaction can be 2 or 1 depending on the interaction type. In case of arity of 1, we can drop the second object.
\blue{Each action in our formulation is realized as a set of object relations in the environment that belongs to $\{\mathrm{OnTop}, \mathrm{Near}, \mathrm{Inside}, \mathrm{ConnectedTo}\}$. The precondition and postconditions of actions are taken directly from prior work~\cite{sharma2022goalnet}. Realization of these conditions using geometric properties is described in Appendix~\ref{sec:manipulation}.}
We shall also use the notion of a timestamp as a subscript to indicate prediction for each
state in the execution sequence.
The space of robot interactions include grasping, releasing, pushing, 
moving an object to another location or inducing discrete state changes 
(e.g.. opening/closing an object, operating a switch or using a mop). 
%
%
We assume the presence of an underlying 
low-level metric planner, encapsulated as a robot \emph{skill}, which  
realizes each symbolic action or returns if the action is infeasible. 
Robot actions are stochastic, modeling 
execution errors (unexpected collisions) and unanticipated outcomes 
(objects falling, changing the symbolic state).
Let $\mathcal{T}(\cdot)$ denote the transition function.  
The successor state $s_{t+1}$ upon 
taking the action $a_{t}$ in state $s_{t}$ is \blue{generated} 
from a physics simulator. 
Let $\eta_{t} = \{ a_{0}, a_{1}, \dots, a_{t-1} \}$ denote the 
\emph{action history} till time $t$.

\subsection{Semantic Goals and Interactions} 
The robot's goal is to perform tasks such as transporting or delivering 
objects to appropriate destinations, making an assembly, clearing or packing items or performing 
abstract tasks such as illuminating or cleaning the room. 
The robot is instructed by providing a \emph{declarative} goal 
$g$ expressing the symbolic constraint between world objects~\cite{hubner2006programming}.
For example, the \emph{declarative} goal, ``place milk in fridge'' 
can be expressed as a constraint $\mathrm{Inside(milk_{0}, fridge_{0})}$ 
between specific object instances.
There may be multiple instance of the same object, for eg. milk carton, in the environment. The sub-index is used to specify which instance is being referred to.
Another example is of the task of moving all fruits onto the kitchen table 
that can be expressed as a list of constraints $\mathrm{OnTop(apple_{0}, table_{0})}$, $\mathrm{OnTop(orange_{0}, table_{0})}$ and $\mathrm{OnTop(banana_{0}, table_{0})}$.
Finally, the robot must synthesize a plan to satisfy the goal constraints.               
Goal-reaching plans may require using some objects 
as tools, for instance, using a container for moving items, 
or a ramp negotiate an elevation. 
\subsection{Predicting Tool Interactions}  
We assume that the robot is primed with a set of primitive symbolic actions 
but lacks knowledge about how object characteristics can facilitate their use 
as in attaining high-level goals. 
Hence, the robot cannot predict
the use of tray-like objects in transportation tasks, or the use of a stick to fetch an object at a distance. 
An exception is by discovering such characteristics via explicit simulation, which may be infeasible or intractable in large planning domains.
Our goal is to learn common sense knowledge 
about \emph{when} an object can be used as a tool 
and \emph{how} their use can be sequenced for goal-reaching plans.
%
We aim at learning a policy $\pi$ 
that estimates the next action $a_{t}$ conditioned on the 
the goal $g$ and the initial state $s$ (including the action history $\eta_{t}$, 
such that the robot's \emph{goal-reaching} likelihood is maximized.  
We adopt the \textrm{MAXPROB-MDP} \cite{kolobov2012planning} formalism and 
estimate a policy that maximizes the goal-reaching likelihood from the given state. 
\textrm{MAXPROB-MDP} can be equivalently viewed as 
an infinite horizon, un-discounted MDP with a zero reward for non-goal states and a 
positive reward for goal states \cite{kolobov-icaps11}.
Formally, let $P^{\pi}\left( s, g \right) $ denote the \emph{goal-probability} function  
that represents the likelihood of reaching the goal $g$ from a state $s$ on 
following $\pi$.
Let $S^{\pi_{s}}_{t}$ be a random variable denoting the state resulting from 
executing the policy $\pi$ from state $s$ for $t$ time steps. 
Let $\mathcal{G}(s,g)$ denote the Boolean \emph{goal check} function that determines  
if the intended goal $g$ is entailed by a world state $s$ as 
$\mathcal{G}(s,g) \in \{\mathrm{True(T)},\mathrm{False(F)} \}$. 
%
The policy learning objective is formulated as 
maximizing the likelihood of reaching a goal-satisfying state $g$ 
from an initial state $s_{0}$, denoted as
\begin{gather}
    \max_{\pi} P^{\pi}(s_{0}, g)=
    \max_{\pi} 
    \sum_{t=1}^{\infty}  P \bigg( \mathcal{G} ( S^{\pi_{s_{0}}}_{t}, g )= \mathrm{T} :  \mathcal{G}( S^{\pi_{s_{0}}}_{t'}, g) = \mathrm{F},   
    ~\forall t' \in [1,t)\bigg). 
\end{gather}
%
 
%
The policy is modeled as a function $f_{\theta}(.)$ parameterized by parameters 
$\theta$ that determines the next action for a given world state, the robot's action history and the 
goal as $a_{t} = f_{\theta} \left( s_{t}, g, \eta_{t} \right) $. 
We adopt imitation learning approach and learn the function $f_{\theta}(.)$ from
demonstrations by human teachers. 
Let $\mathcal{D}_{\mathrm{Train}}$ denote the corpus of $N$ goal-reaching plans, 
\begin{equation}
    \mathcal{D_{\mathrm{Train}}} = \{ (s_0^i,g^i,\{s_j^i,a_j^i\}) \mid i \in \{1, \ldots, N\}, j \in \{0,t_{i}-1\} \}, 
    \label{eq:dataset}
\end{equation}
where the $i^{th}$ datum 
consists of the initial state $s^{i}_0$, the goal $g^{i}$ and a state-action sequence 
\[\{ (s^{i}_{0}, a^{i}_{0}), \dots, (s^{i}_{t-1}, a^{i}_{t-1}) \}\] of length $t_{i}$. 
The set of human demonstrations elucidate common sense knowledge 
about \emph{when} and \emph{how} tools can be used for attaining provided goals.  
The data set $\mathcal{D}_{\mathrm{Train}}$ supervises an imitation loss 
between the human demonstrations and the model predictions,
resulting in learned parameters $\theta^{*}$.  
%
 Online, the robot uses the learned model to sequentially predict actions and execute in the 
simulation environment till the goal state is attained. 
%
We also consider the \emph{open-world} case where the robot may 
encounter instances of \emph{novel} object categories \emph{unseen} in training, 
necessitating a \emph{zero-shot} generalization.  

%% file: technical_approach.tex
\section{Technical Approach}\label{sec:model}





We aim at predicting the next robot action $a_{t}$,  
given the world state $s_{t}$, the intended goal $g$ and the 
history $\eta_{t}$ of past actions taken the robot. 
We consider learning to predict multi-step plans requiring complex interaction of objects as tools in possibly novel environments unseen during training. 
Our technical approach is built on three insights. First, we factor the overall learning task of predicting the action for each environment state by first predicting the tool required to perform the task and then predicting required interaction. This tool conditioned action prediction enables us to decouple the learning problem and make the model training tractable.  
Second, we incorporate learning dense embeddings of the robot's environment as well as semantic representations for words trained on existing symbolic knowledge corpora. The learned representation allows the robot to generalize its predictions to novel environments populated with novel objects that may share semantic attributes with those encountered during training. 
Finally, we turn to a corpus of human demonstrated plans to train our learner. The corpus elucidates the commmon sense knowledge of sequencing tools interactions for a task.  
Formally, we introduce an imitation learner, realized as a 
hyper-parametric function (a neural network model) denoted as $f_{\theta}$ as follows: 

\begin{equation}
  a_{t} = f_{\theta} \left( s_{t}, g, \eta_{t} \right) 
 = 
 f^{act}_{\theta}  \left( f^{goal}_{\theta} \left( f^{state}_{\theta} \left( s_{t} \right), g,  f^{hist}_{\theta} \left( \eta_{t} \right) \right) , f^{tool}_{\theta}(s_t, g, \eta_t) \right).
\label{eqn:nn}
\end{equation}
It adopts an object-centric graph representation, learning 
a state encoding that fuses metric-semantic information about 
objects in the environment via function $f^{state}_{\theta}\left(\cdot \right)$. 
The function $f^{hist}_{\theta} \left(\cdot \right)$ encodes the action history. 
\blue{The model learns to attend over the world state conditioned 
on the declarative goal and the history of past actions through $f^{goal}_{\theta}\left( \cdot \right)$.  }
We leverage an adapted version of our tool likelihood prediction model \textsc{ToolNet}, denoted as $f^{tool}_\theta\left(\cdot \right)$.
Finally, the learned encodings are decoded as the next action for the 
robot to execute via $f^{act}_{\theta}\left( \cdot \right)$.   
Crucially, the predicted action is grounded over an 
\emph{a-priori} unknown state and type of objects in the environment. 
The predicted action is executed in the environment and the  updated state action history is used for estimation 
at the next time step. Using a neural network as a hyper-parametric function for action prediction enables us to leverage ConceptNet embeddings to generalize and scale with the size of object sets, goal and interaction types. We utilize an imitation learning model with the dataset described in Equation~\ref{eq:dataset} to learn the $\theta$ parameters of the neural network.
The constituent model components are detailed next. 
%

\vspace{0.5ex}
\noindent 
\subsection{Graph Structured World Representation }
%

\begin{figure*}[t]
    \centering
    \setlength{\belowcaptionskip}{-10pt}
    \includegraphics[width=\textwidth]{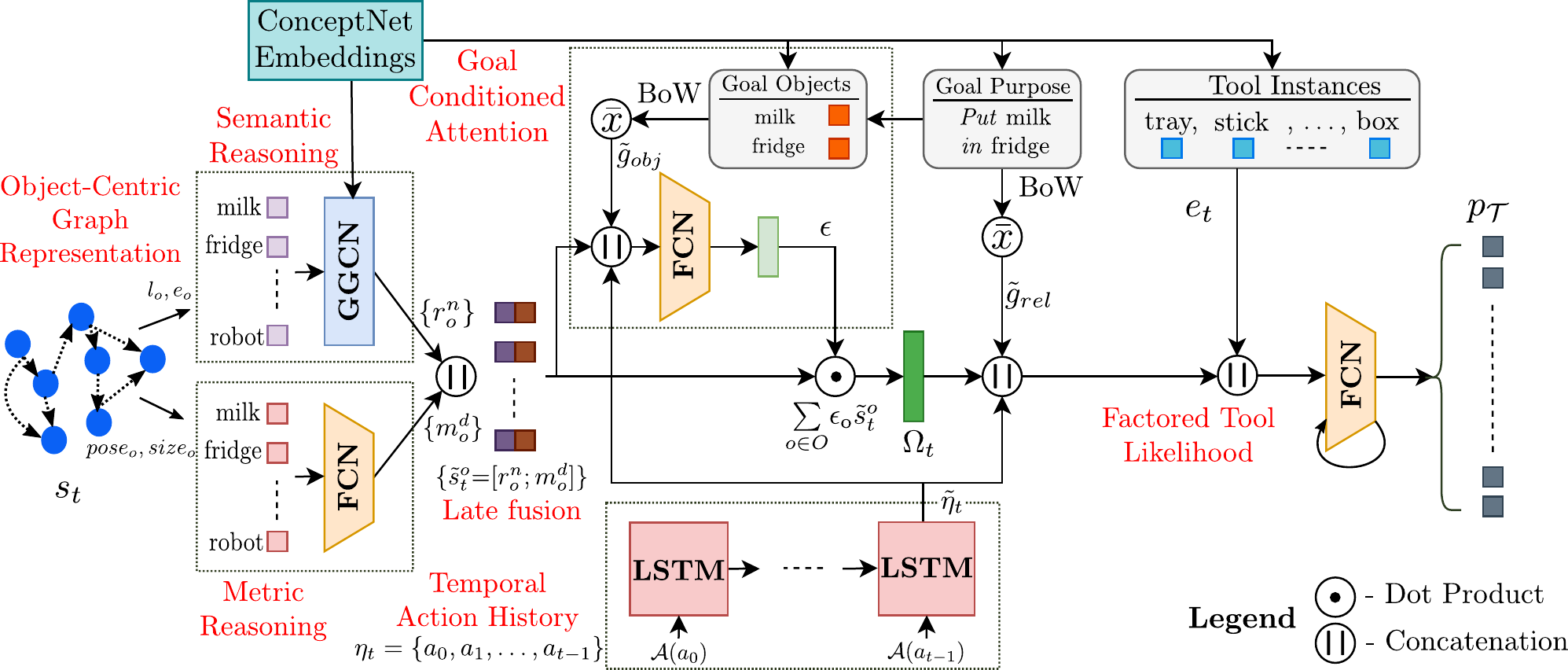}
  \caption{ 
  Updated \textsc{ToolNet} model encodes the metric-semantic world state using graph convolution (GGCN) and fully connected (FCN) layers. The model uses goal information and the robot's action history to attend over a task-specific context, finally predicting the likelihood scores for each tool in the environment. A graph structured representation and inclusion of pre-trained word embeddings (from a knowledge base) facilitate generalization in predicting interactions in novel contexts with new objects unseen in training. 
  }  
    \label{fig:toolnet2}
\end{figure*}

\subsubsection{Semantic Reasoning}
\label{sec:semantic}
We first describe the $f^{state}_\theta\left(\cdot \right)$ function in equation \eqref{eqn:nn}. We denote the robot's current world state $s_t$  as an object-centric graph 
$G_t = (O, R)$. 
Each node in the graph represents an object instance $o \in O = \mathcal{O}(s_t)$.
The edge set consists of binary relationships $\mathrm{OnTop}$, $\mathrm{ConnectedTo}$, $\mathrm{Near}$ and $\mathrm{Inside}$ between objects $R \subseteq O \times O$.  
%
%
Let $l_{o} \in \{0,1\}^{p}$ represents discrete object states for the object $o$ (e.g. $\mathrm{Open/Closed}$, $\mathrm{On/Off}$). Here, $p$ represents the number of possible states of all objects.
Next, it incorporates a pre-trained function $\mathcal{C}(\cdot)$ that embeds a 
word (like token of an object class or a relation) to a dense distributed representation, such that semantically close 
tokens appear close in the learned space \cite{mikolov2018advances}. 
The use of such embeddings enables generalization, which we discuss subsequently. Consider the example goal of ``Place the box in the cupboard''. In this case we want to satisfy the relationship $\mathrm{Inside}$ for cupboard and box. The entire environment can be considered as the graph where relations like $\mathrm{OnTop}(box, table)$ may be true. The pretrained function $\mathcal{C}(\cdot)$ can be any pretrained language model.

%
Let $e_{o} = \mathcal{C}(o) \in \mathcal{R}^{q}$ denote the $q$-dimensional embedding for an object instance $o$. 
The embeddings $l_{o}$ and $e_{o}$ model object attributes that initialize the 
state of each object node in the graph. 
%
%
The local context for each $o$ is incorporated via 
a Gated Graph Convolution Network (GGCN)  \cite{liao2019synthesizing}, which
performs message passing between 1-hop vertices on the graph. 
Following \cite{puig2018virtualhome}, the gating stage is realized 
as a Gated Recurrent Unit (GRU) resulting in \emph{graph-to-graph} 
updates as: 
\begin{align}
\begin{split}
    r_o^{0} &= \mathrm{tanh} \left (W_{r} \left[ l_{o}; e_{o} \right] + b^r \right),\\[-3pt]
    x^k_o &= \sum_{j \in R}\sum_{o' \in N^j(o)} W_j^k r_{o'}^{k-1} ,\\[-3pt]
    r^k_{o} &= \mathrm{GRU} \left( r^{k-1}_o, x^{k}_{o} \right).\\[-3pt]
\end{split}
\end{align}
Here, the messages for object $o$ are aggregated over neighbors $N^j(o)$ connected by relation $j$ ($\forall j \in R$) 
during $n$ convolutions, resulting in an embedding $r^{n}_{o}$ for each object instance in the environment. Unlike a fully-connected neural network, a GGCN model facilitates inference over the relations across objects for an input object centric graph. These relations are crucial to predict the relevant actions and achieve the intended goal. GRUs in the convolution iterations facilitate stable training of the model~\cite{li2015gated}.
%
%

\subsubsection{Fusing Metric Information. }
Next, we incorporate the metric information associated with objects. 
Let $pose_o$ and $size_o$ represent the pose and size/extent (along xyz axes) 
for each object instance. \blue{The pose} includes the spatial position and the orientation of the object. We do not explicitly provide the point-cloud geometric model and abstract out the surface information that may be required to infer the utility of objects as tools and instead rely on ConceptNet embeddings for the same. However, in our evaluation (Section~\ref{sec:experiments}), we utilize CAD models that closely represent the surface properties to model the physics of tool use. Another point to note here is that for the specific task of utilizing an object as a tool the pose information has little contribution. However, the pose information is required to accurately identify the system state. For instance, the pose information of the door of a fridge indicates whether the fridge is open or closed. This information is given as a categorical value \textit{i.e.} the state vector $l_o$ to the GGCN model. To explicitly pass this information to the FCN model, we use the pose vector of objects as inputs.
The properties are encoded using a $d$-layer Fully Connected Network (FCN) with a 
Parameterized ReLU (PReLU) \cite{prelu} activation as: 
\begin{align}
\begin{split}
	m^0_o &= \mathrm{PReLU} \left( W_{mtr}^{0}[pose_o; size_o] + b_{mtr}^0\right) \\[-3pt]
    m^k_o &= \mathrm{PReLU} \left( W_{mtr}^{k} m^{k-1}_o  +b_{mtr}^{k}\right), 
\end{split} 
\end{align}
resulting in the metric encoding $m^d_o$ for each object in the scene. 
A world state encoding (for~$s_t$) is obtained by fusing 
the semantic and metric embeddings as 
\begin{equation}
\label{eq:state}
    f^{state}_{\theta}(s_t) = \{\tilde{s}_t^o\! =\! [r^n_o;m^d_o]|\ \forall o \in {\cal O}(s_t)\}.
\end{equation}
\emph{Late fusion} of the two encodings 
allows downstream predictors to exploit them independently.
 
\vspace{0.5ex}
\noindent 
\subsection{Encoding the Action History }
Next, we define the $f^{hist}_\theta\left(\cdot \right)$ function in equation \eqref{eqn:nn}.
The task of deciding the next action is informed by the agent's action history in two ways.
First, sequential actions are often temporally correlated. 
For example, a placing task often involves moving close to the box, 
opening it and then placing an item inside it. \blue{ If the previous action involved moving close to a box, this information facilitates the model to leverage localized contextual information and not jump to manipulate another object in the goal specification.} This, in tandem with the goal-conditioned attention facilitates seamless action sequence generation. 
Hence, maintaining the action history can help in prediction of the next action. 
\blue{Second, the set of actions the robot executed in the past 
provides a local context indicating the objects the 
the robot may utilize in the future.}
Formally, we encode the temporal action history $\eta_t$
using an $\mathrm{LSTM}$. 
We define action encoding $\mathcal{A}(a_{t-1})$ of 
$a_{t-1} = I_t (o^1_{t-1}, o^2_{t-1})$ independent of the object set, as 
$\mathcal{A}(a_{t-1}) = [\vec{I}_{t-1}; \mathcal{C}(o^1_{t-1}); \mathcal{C}(o^2_{t-1})]$, 
where $\vec{I}_{t-1}$ is a one-hot vector over possible interaction types 
$\mathcal{I}$, and 
$\mathcal{C}(o^1_{t-1})$ and $\mathcal{C}(o^2_{t-1})$ 
represent the word embeddings of the 
object instances $o^1_{t-1}$ and $o^2_{t-1}$. 
At each time step $t$, the $\mathrm{LSTM}$ 
encoder takes in the encoding of previous action, $\mathcal{A}(a_{t-1})$ and 
outputs the updated encoding $\tilde{\eta}_t$, given as  
$\tilde{\eta}_t = \mathrm{LSTM}(\mathcal{A}(a_{t-1}),\tilde{\eta}_{t-1})$.
%
This results in the embedding vector 
\begin{equation}
\label{eq:history}
    f^{hist}_{\theta}(\eta_t) = \tilde{\eta}_t.
\end{equation}

\subsection{Goal-conditioned Attention}
We now define the $f^{goal}_\theta\left(\cdot \right)$ function in equation \eqref{eqn:nn}.
The goal $g$ consists of symbolic relations (e.g. $\mathrm{Inside}$, $\mathrm{OnTop}$ etc.) 
between object instances (e.g., carton and cupboard) that must be true at the end of the robot's plan execution. 
The declarative goal input to the model is partitioned as relations 
$g_{rel}$ and the object 
instances specified in the goal $g_{obj}$.
The resulting encondings are denoted as $\tilde{g}_{rel}$ and $\tilde{g}_{obj}$:
\begin{equation*}
    \tilde{g}_{rel} = \frac{1}{|g_{rel}|}  \sum_{j \in g_{rel}} \mathcal{C}(j) 
   \hspace{2mm} \mathrm{ and }    \hspace{2mm}
    \tilde{g}_{obj} = \frac{1}{|g_{obj}|} \sum_{o \in g_{obj}} \mathcal{C}(o).
\end{equation*}
Next, the goal encoding and the 
action history encoding $\tilde{\eta}_{t}$ is used to learn attention weights over objects in the 
environment \cite{bahdanau2014neural} such that 
\begin{align}
\begin{split}
    \epsilon_o = \mathrm{softmax} \left( W_g [ \tilde{s}_t^o; \tilde{g}_{obj}; \tilde{\eta}_t ]  + b_g \right),
\end{split}
\end{align}
where $\tilde{s}_t^o$ is obtained from $f^{state}_\theta(s_t)$ in~\eqref{eq:state} and $\tilde{\eta}_t$  is obtained from $f^{hist}_{\theta}(\eta_t)$ in~\eqref{eq:history}. 
This results in the attended scene encoding $\Omega_{t}$ as:
\begin{equation}
\label{eqn:attn_scene}
    \Omega_{t} = f^{goal}_\theta(\tilde{s}_{t}^o, \tilde{g}_{obj}, \tilde{\eta}_{t}) =  \sum_{o \in O} \mathrm{\epsilon_o} \tilde{s}_t^o. 
\end{equation}
The attention mechanism aligns the goal information with the scene learning a task-relevant context,  
relieving the model from reasoning about objects in the environment unrelated to the task, 
which may be numerous in realistic environments. 

\subsection{Tool Likelihood Prediction}
Now we define the function that predicts the likelihood scores for each tool in the environment state, \textit{i.e.}, the $f^{tool}_\theta\left(\cdot \right)$ function in equation \eqref{eqn:nn}. This likelihood score corresponds to the probability that a particular tool could be used to reach a goal state for a given environment state and declarative goal specification.
In order to allow the model to generalize to unseen tools, instead of prediction over the pre-defined tool set $\hat{\tau}$, we allow the model to predict a likelihood score of a tool $t$ (which may not be present in any of the scenes in the training set) to be used as a tool using the object embedding ($e_t = \mathcal{C}(t)$ as described in Section~\ref{sec:semantic}). This recurrence is shown in the factored tool likelihood module in Figure~\ref{fig:toolnet2}. The prediction is made using the encoding of the state, i.e the attended scene embedding $\Omega_t$, the relational description of the goal $\tilde{g}_{rel}$ and the action history encoding $\tilde{\eta}_t$. Likelihood of each tool is computed for each $t$ using,
\begin{equation}
    p_{t} = \mathrm{sigmoid}(W[\Omega_t; \tilde{g}_{rel}; \tilde{\eta}_t; e_t])\ \forall\ t \in \hat{\tau},
\end{equation}
where $\Omega_t$ is obtained using~\eqref{eqn:attn_scene} and $\tilde{\eta}_t$  is obtained using~\eqref{eq:history}.
We now predict over a tool set $\tau$, which may be different from the fixed toolset seen during training set. \blue{We include the possibility of having tools present at inference time that are unseen during training. The factored style likelihood prediction enables our model to be flexible to make likelihood predictions for unseen tools.}
We now use this to define the likelihood score for each object as $p_t$ for tools and $0$ otherwise. \blue{Here $p_t$ is the probability that tool t can be used to complete the goal.} Specifically, 
\begin{equation}
\label{eq:tool}
    f^{tool}_{\theta}(s_t, g_t, \eta_t) = \{p_o \text{ if } o \in \tau \text{ else } 0\ |\ \forall o \in {\cal O}(s_t)\}.
\end{equation}
Unlike the original \textsc{ToolNet} model~\cite{toolnet}, $f^{tool}_\theta\left(\cdot \right)$ function in this work can predict the tool likelihood scores for each time-step $t$ and utilizes action history encoding $\tilde{\eta}_t$ to capture temporal context.

\begin{figure*}[!t]
    \centering
    \setlength{\belowcaptionskip}{-10pt}
    \includegraphics[width=\textwidth]{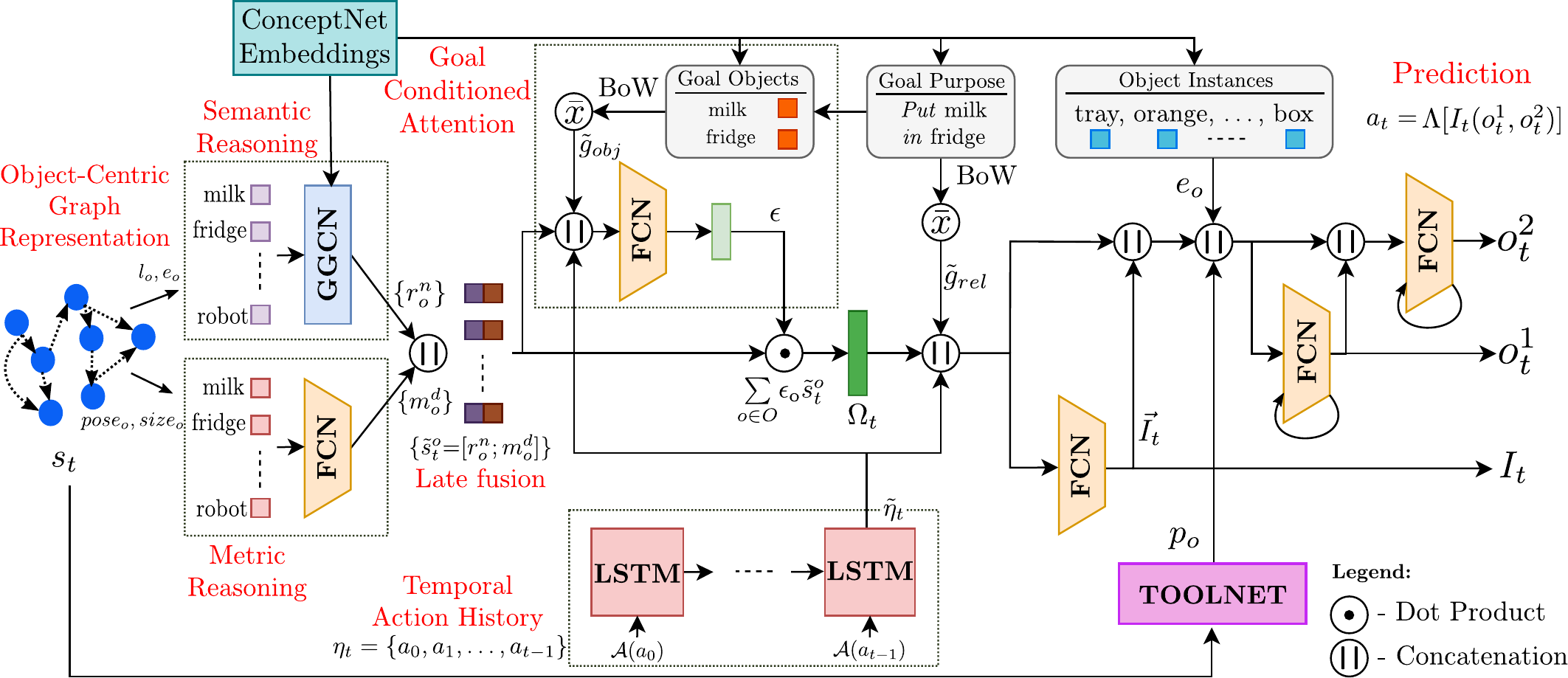}
  \caption{ 
  \modelname neural model is architecturally similar to \textsc{ToolNet} but also incorporates tool likelihood scores predicted from the \textsc{ToolNet} model, finally decoding the next symbolic action for the robot to execute. The interaction type and objects are predicted auto-regressively and in a factored style. 
  }  
    \label{fig:complete_model}
\end{figure*}

\subsection{Robot Action Prediction} 
We now take the encoded information about the world state, goal and action history to decode the next symbolic action $a_t = I_t (o^1_t, o^2_t)$.
%
The three components $I_t$, $o^1_t$ and $o^2_t$ are predicted auto-regressively, where the prediction of the interaction, $I_t$ is used for the prediction of the first object, $o^1_t$ and both their predictions are used for the second object prediction, $o^2_t$.
For the object predictors $o^1_t$ and $o^2_t$, instead of predicting over a predefined set of objects, our model predicts a likelihood score of each object $o\in O$ based on its object embedding $ \tilde{s}_t^o$ and tool likelihood score $p_o$ from~\eqref{eq:tool}. It then selects the object with highest likelihood score. 
The resulting factored likelihood allows the model to generalize to an \emph{a-priori} unknown number and types of object instances:   
%
\begin{gather}
    I_t = \mathrm{argmax}_{I \in \mathcal{I}} \left( \mathrm{softmax}(W_I [\Omega_t; \tilde{g}_{rel}; \tilde{\eta}_t] + b_I) \right),\\
    \begin{split}
    \label{eq:o1}
    o^1_t &= \mathrm{argmax}_{o \in {O}} \alpha_t^o \\ &= \mathrm{argmax}_{o \in {O}}\ ( \sigma(W_{\alpha} [\Omega_t; \tilde{g}_{rel}; \tilde{\eta}_t; e_o; p_o; \vec{I}_t] + b_{\alpha})) \mathrm{,}
    \end{split}\\
    \label{eq:o2}
    o^2_t = \mathrm{argmax}_{o \in {O}}\ ( \sigma(W_{\beta} [\Omega_t; \tilde{g}_{rel}; \tilde{\eta}_t; e_o; p_o; \vec{I}_t; \alpha_t^o] + b_{\beta})).
\end{gather}
Here $\alpha^o_t$ denotes the likelihood prediction of the first object. 
%
Finally, we impose grammar constraints (denoted as $\Lambda$) at inference time 
based on the number of arguments that the predicted interaction $I_t$ accepts. 
If $I_t$ accepts only one argument, only $o^1_t$ is selected, otherwise 
both are used. 
%
%
Thus, predicted action for time-step $t$ is denoted as
\begin{equation}
    a_t = f^{act}_{\theta}(\Omega_t,\tilde{g}_{rel},\tilde{\eta}_t) = \Lambda [ I_t (o^1_t, o^2_t) ].
\end{equation}
This action is then executed by the robot in simulation. Our simulation model follows closely to the one developed by~\citeA{puig2018virtualhome} wherein symbolic actions such as $\mathrm{moveTo}$, $\mathrm{pick}$ and $\mathrm{place}$ are realized as relations in the environment state using a PyBullet simulator. For instance, the $\mathrm{moveTo}$ action changes the xyz coordinates of the robot such that it is $\mathrm{Near}$ the target object. \blue{Similarly,} $\mathrm{pick}$ establishes a $\mathrm{connectedTo}$ relation between the target object and robot arm. For implementation specific details, visit \url{https://github.com/reail-iitd/tango/wiki}.
The executed action and resulting world state is provided as input to the model 
for predicting the action at the next time step. 
The $f^{act}_{\theta}\left(\cdot \right)$ function denotes the \textsc{ToolTango} model. In our original \textsc{Tango} model presented  in~\cite{tango}, the $p_o$ tool likelihood score was not part of equations~\eqref{eq:o1} and~\eqref{eq:o2}.
 

\subsection{Word Embeddings Informed by a Knowledge Base}
\modelname uses word embedding function $\mathcal{C}(\cdot)$ that provides a  
dense vector representation for word tokens associated with object class and relation types.  
%
%
Contemporary models
use word embeddings acquired from language modeling tasks \cite{mikolov2018advances}. 
We adopt embeddings that are additionally informed by an existing knowledge graph 
$\mathrm{ConceptNet}$ \cite{speer2017conceptnet} that 
 provides a sufficiently large knowledge graph connecting words with edges expressing relationships such as 
$\mathrm{SimilarTo}$, $\mathrm{IsA}$, $\mathrm{UsedFor}$, $\mathrm{PartOf}$ and $\mathrm{CapableOf}$. 
Word embeddings \cite{mikolov2018advances} can be \emph{retro-fitted} 
such that words related using knowledge graph embeddings are also close in the embedding space \cite{conceptnet-github}. 
Using such (pre-trained) embeddings incorporates \emph{general purpose} relational 
knowledge to facilitate richer generalization for downstream policy learning. Thus, given an object, say $apple$, the word embedding function $\mathcal{C}(.)$ returns a dense $\mathrm{ConceptNet}$ \blue{vector} that corresponds to this token.
The complete sequence of steps is summarized in Figure~\ref{fig:complete_model}.

\subsection{Model Training}

We decompose the action prediction task into first predicting tool, \textit{i.e.}, the $f^{tool}_\theta\left(\cdot \right)$ function and then evaluating the $f^{act}_\theta\left(\cdot \right)$ function. We also decompose model training. First, we train the tool prediction model. The loss used to train this model is Binary Cross-Entropy with each $t \in \hat{\tau}$ acting as a class. The class label, $y^i_t$ is assigned 1 if it is used in a given demonstration, $i$ and 0 otherwise. We also use categorical weights based on plan execution time to encourage shorter plans~\cite{toolnet}. However, the knowledge of the time taken for different plans has not been injected into the model.
In order to make this notion explicit to the model we use loss weighting  such that for the dataset $\mathcal{D_{\mathrm{Train}}}$ defined in~\eqref{eq:dataset},
\begin{equation}
    \mathcal{L} = - \sum_i \alpha_i \sum_j \sum_{t \in \tau} y^i_{j,t} \log(p^i_{j,t}) + (1 - y^i_{j,t}) \log(1 - p^i_{j,t}),
\end{equation}
where, $p^i_{j,t}$ is obtained from $f^{tool}_\theta(s^i_j, g^i, \eta^i_j)$ for each datapoint in $\mathcal{D_{\mathrm{Train}}}$ and $\alpha_i$ is a multiplier that is high for optimal plans (shortest among human demonstrations) and low otherwise.

For a trained tool prediction model, we then train the $f^{act}_\theta\left(\cdot \right)$ \modelname model, fine-tuning the weights of our updated \textsc{ToolNet} function $f^{tool}_\theta\left(\cdot \right)$. Here too we use the Binary Cross-Entropy loss, with the loss for the three predictors (action and the two objects) being added independently. Additional model training and hyperparameter details are given in Appendix~\ref{appendix:hyperparams}.

\blue{This decomposed training style has direct implications on the training stability. First training \textsc{ToolNet} to predict the likelihood scores for tools enables action prediction informed by the likelihood scores of each tool in the environment. }

%% file: data_collection.tex
\section{Data Collection Platform and Annotation}
\label{sec:data}
\begin{table*}[]
    \centering
    \resizebox{\textwidth}{!}{
    \begin{tabular}{|c|c|c|c|c|c|}
    \hline 
    \multirow{2}{*}{Domain} & \multirow{2}{*}{Plan lengths} & Objects interact- & Tools used & \multirow{2}{*}{Sample objects} & \multirow{2}{*}{Sample goal specifications}\tabularnewline
     &  & ed with in a plan & in a plan &  & \tabularnewline
    \hline 
    \hline 
    Home & 23.25$\pm$12.65 & 4.12$\pm$1.97 & 0.93$\pm$0.70 & \begin{minipage}[t]{0.48\textwidth}floor$^{1}$, wall, fridge$^{123}$, cupboard$^{123}$, tables$^{1}$, couch$^{1}$, \textbf{big-tray}$^{1}$, \textbf{tray}$^{1}$, \textbf{book}$^{1}$,
        paper, cubes, light switch$^{4}$, bottle, \textbf{box}$^{2}$, fruits, \textbf{chair}$^{15}$, \textbf{stick}, dumpster$^{2}$, milk carton, shelf$^{1}$, \textbf{glue}$^{6}$, \textbf{tape}$^{6}$, \textbf{stool}$^{15}$, \textbf{mop}$^{8}$, \textbf{sponge}$^{8}$, \textbf{vacuum}$^{8}$, dirt$^{7}$, door$^{2}$\end{minipage} & \begin{minipage}[t]{0.32\textwidth} 1. Place milk in fridge, 2. Place fruits in cupboard, 3. Remove dirt from floor, 4. Stick paper to wall, 5. Put cubes in box, 6. Place bottles in dumpster, 7. Place a weight on paper, 8. Illuminate the room. \vspace{2pt}
        \end{minipage}\tabularnewline
    \hline 
    Factory & 38.77$\pm$23.17 & 4.38$\pm$1.85 & 1.44$\pm$0.97 & \begin{minipage}[t]{0.48\textwidth}
        floor$^{1}$, wall, \textbf{ramp}, worktable$^{1}$, \textbf{tray}$^{1}$, \textbf{box}$^{2}$, crates$^{1}$, \textbf{stick}, long-shelf$^{1}$, \textbf{lift}$^{1}$, cupboard$^{123}$, \textbf{drill}$^{4}$, \textbf{hammer}$^{49}$, \textbf{ladder}$^{5}$, \textbf{trolley}$^{2}$, \textbf{brick}, \textbf{blow dryer}$^{48}$,
        \textbf{spraypaint}$^{4}$, \textbf{welder}$^{4}$, generator$^{4}$, \textbf{gasoline}, \textbf{coal}, \textbf{toolbox}$^{2}$, \textbf{wood cutter}$^{4}$, \textbf{3D printer}$^{4}$, screw$^{9}$, nail$^{9}$, \textbf{screwdriver}$^{49}$, wood, platform$^{1}$, oil$^{7}$, water$^{7}$,
        board, \textbf{mop}$^{8}$, \textbf{glue}$^{6}$, \textbf{tape}$^{6}$, \textbf{stool}$^{15}$\vspace{2pt}\end{minipage} & \begin{minipage}[t]{0.32\textwidth} 1. Stack crated on platform, 2. Stick paper to wall, 3. Fix board on wall, 4. Turn on the generator, 5. Assemble and paint parts, 6. Move tools to workbench, 7. Clean spilled water, 8. Clean spilled oil. \vspace{2pt}
        \end{minipage}\tabularnewline
    \hline 
    \end{tabular}} \setlength{\belowcaptionskip}{-6pt}
    \caption{{Dataset characteristics. The average plan length measured as the length of action sequence), number of objects interacted in plan and number of tools used in plans with object and goal sets for Home and Factory domains. Object positions were sampled using Gaussian distribution. Objects in bold can be used as tools. Legend:- $^{1}$:~surface, $^{2}$:~can open/close, $^{3}$:~container, $^{4}$:~can operate, $^{5}$:~can climb, $^{6}$:~can apply, $^{7}$:~can be cleaned, $^{8}$:~cleaning agent, $^{9}$:~can 3D print. Objects in bold can be used as tools. Object affordances derived from properties in the ConceptNet graph~\cite{speer2017conceptnet}.} Stool/ladder are objects used to represent a tool for raising the height of the robot.}
    \label{tab:dataset_complete}
\end{table*}

\subsection{Data Collection Environment} 
We develop a low-fidelity simulation environment where the robot can take actions and interact with objects. The low-level physics of motion is less important as we are primarily concerned with the symbolic effects on the world state. 
To do this, we use PyBullet, a physics simulator~\cite{coumans2016pybullet}, 
to generate home and  factory-like environments  
populated with a virtual mobile manipulator (a Universal Robotics (UR5) arm 
mounted on a Clearpath Husky mobile base).
The world state is object-centric including the metric locations of 
objects and the discrete states of symbolic attributes and relations. 
The objects in the domains were derived from real-world home and factory scenes that span Facebook Replica Dataset \cite{straub2019replica}. These scenes were made to look as photo realistic as possible. The objects on other hand were chosen to span the YCB dataset \cite{calli2017ycb}. The CAD model for each object obtained from open-source repositories such as the Google 3D Warehouse\footnote{~\url{https://3dwarehouse.sketchup.com/}}. This was done in order to keep the simulated physics as real as possible. The set of objects in the two domains are listed in Table~\ref{tab:dataset_complete}.
 
Each object in the environment is associated with a metric location and physical extent and may 
optionally possess discrete states such as $\mathrm{Open/Closed}$, $\mathrm{On/Off}$, etc. 
The world model is assumed to possess spatial notions 
such as $\mathrm{Near}$ or $\mathrm{Far}$. 
Objects in the world model can be supported by, contained within or 
connected with other objects (or the agent). Hence, we include  
semantic relations such as $\mathrm{OnTop}$, $\mathrm{Inside}$, $\mathrm{ConnectedTo}$ etc. More details in Appendix~\ref{sec:manipulation}.

\begin{figure}[t]
    \centering
    \resizebox{0.8\columnwidth}{!}{
    \begin{tabular}{|c|}
    \hline 
    \textbf{Robot Actions}\tabularnewline
    \hline 
    \begin{minipage}[t]{\columnwidth}Push, Climb up/down, Open/Close, Switch on/off, Drop, Pick, Move to, Operate device,
    Clean, Release material on surface, Push until force\end{minipage}\tabularnewline
    \hline 
    \hline 
    \textbf{Object Attributes}\tabularnewline
    \hline 
    \begin{minipage}[t]{\columnwidth}Grabbed/Free, Outside/Inside, On/Off, Open/Close, Sticky/Not Sticky, Dirty/Clean, Welded/Not Welded, Drilled/Not Drilled, Driven/Not Driven, Cut/Not Cut, Painted/Not Painted\end{minipage}\tabularnewline
    \hline 
    \hline
    \textbf{Semantic Relations}\tabularnewline
    \hline 
    On top, Inside, Connected to, Near\tabularnewline
    \hline 
    \hline
    \textbf{Metric Properties}\tabularnewline
    \hline 
    Position, Orientation, Size\tabularnewline
    \hline 
    \end{tabular}} \setlength{\belowcaptionskip}{-8pt}
    \captionof{table}{{Domain Representation. Robot symbolic actions, semantic attributes, relations to describe the world state and objects populating the scene in Home and Factory Domains. }}
    \label{tab:env_desc}
\end{figure}

The robot possesses a set of behaviours or symbolic actions such as 
$\mathrm{Moving}$ towards an object, $\mathrm{Grasping}$, $\mathrm{Releasing/Dropping}$ or $\mathrm{Pushing}$ 
an object or  $\mathrm{Operating}$ an entity to imply actions that induce 
discrete state changes such as opening the door before exiting, turning on a switch etc. 
We assume that the robot's actions can be realized by the presence of an 
underlying controller.
We encode the geometric requirements for actions as symbolic pre-conditions. 
Examples include releasing an object from the gripper before grasping another, 
opening the door before trying to exit the room. 
The set of possible actions and the range of interactions  are listed in Table~\ref{tab:env_desc}. 

The robot is tasked with goals that involve multiple interactions with objects  
derived from standardized data sets \cite{calli2017ycb}. 
These goals include: 
(a) \emph{transporting} objects from one region to another (including space on top of or inside
other objects), 
(b) \emph{fetching} objects, which the robot must reach, grasp and return with, and 
(c) \emph{inducing state changes} such as illuminating the room or removing dirt from floor. 
We assume that the robot is instructed by providing \emph{declarative} goals.     
\blue{For example, the task of moving all fruits on top of the kitchen table 
can be modeled as a set intended constraints among the interacting objects.}
The effects of actions such as pushing or moving are simulated via a motion planner and propagated to the next time step. 
Abstract actions such as attachment, operating a tool or grasping/releasing objects are encoded symbolically 
as the establishment or release constraints. {The simulation for these actions is coarse and considers their symbolic effects 
forgoing the exact motion/skills needed to implement them. We assume that the robot can realize abstract actions through low-level routines. }

\begin{figure}
    \centering \setlength{\belowcaptionskip}{-6pt}
    \includegraphics[width=\linewidth]{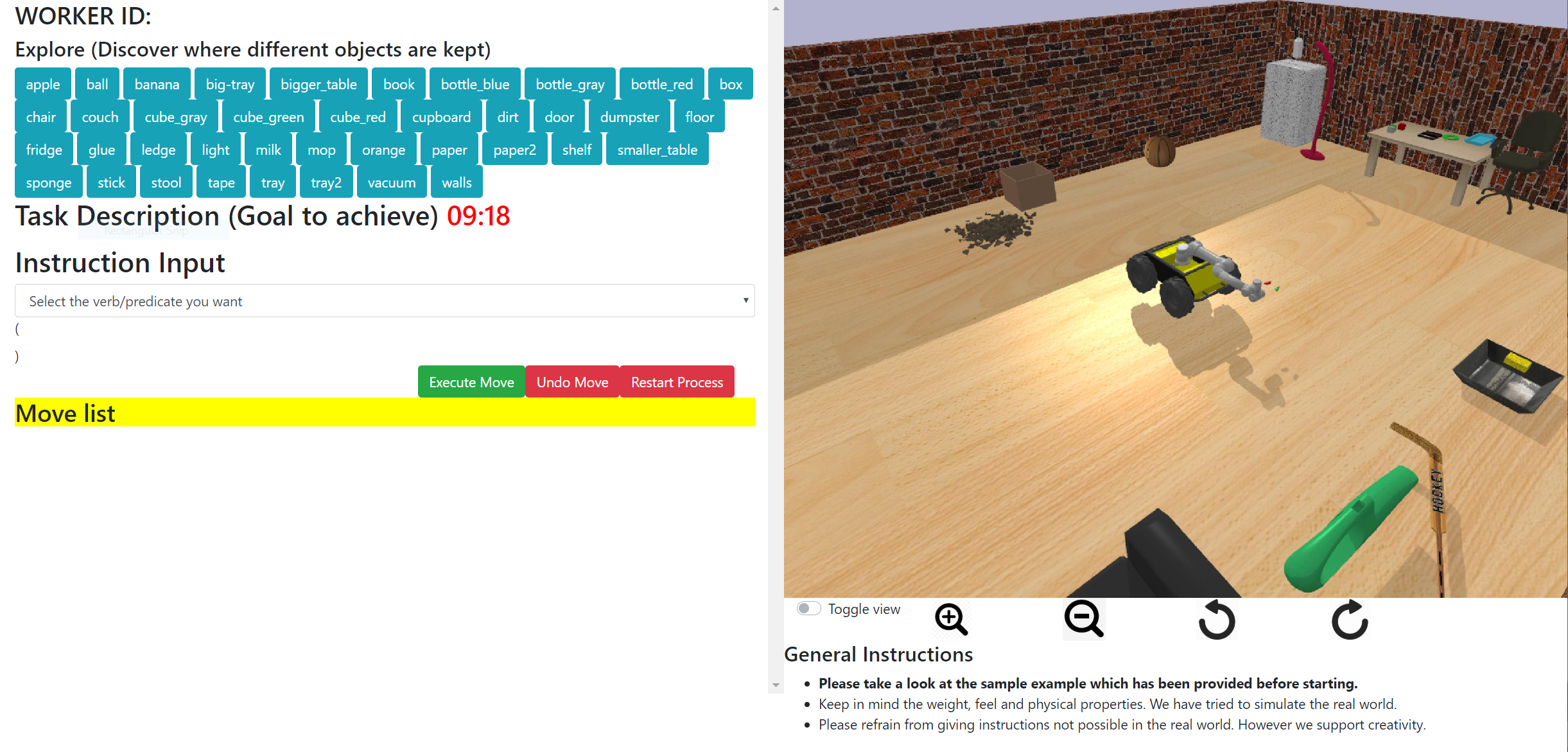}
    \caption{Data Collection Interface. The human teacher instructs (left) a virtual mobile manipulator robot by specifying symbolic actions. The human-instructed plan is simulated and visualized (right). \blue{The user is shown} the goal needed to be completed in text form. They select the user interaction, first object and second object to instruct the robot. The interface can be seen in action in the \protect\url{https://www.youtube.com/watch?v=lUWU3rK1Gno}}.
    \label{fig:screenshot} 
\end{figure}

We present the human agents with a scene and goal conditions we want to attain and receive actions in a custom grammar. The instructions are grounded in the world and the agent and the virtual environment \blue{show} its effects to input the next action. This is repeated till a goal state is reached. Figure~\ref{fig:screenshot} illustrates the interactive platform used for data collection. 
Using a curated dataset, the robot must synthesize a plan of executable actions to 
satisfy the goal constraints. 
The presence of a rich space of interactions gives rise to 
plans with multiple interactions between objects. 
For example, ``packing items into a basket and carrying the basket to the goal region'', 
``using a stick to fetch and drop an object beyond reach into a box'', 
``using a ramp/stool to elevate itself to fetch an object''.

\subsection{Annotated Corpus} 
\begin{figure}[]
    \centering
    \begin{subfigure}{0.48\linewidth}
        \centering
        \includegraphics[width=0.95\linewidth]{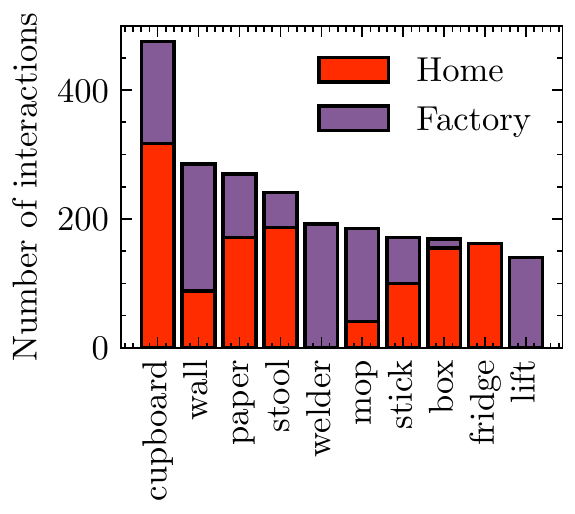}
        \caption{Object interactions} \vspace{-6pt}
        \label{fig:num_int}
    \end{subfigure}
    \begin{subfigure}{0.48\linewidth}
        \centering
        \includegraphics[width=0.95\linewidth]{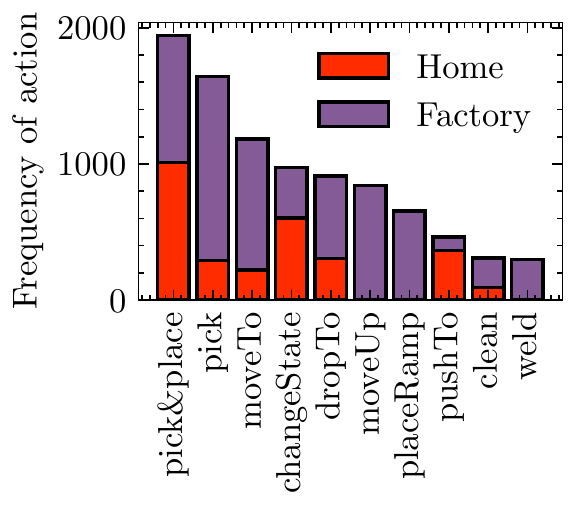}
        \caption{Symbolic actions} \vspace{-6pt}
        \label{fig:num_actions}
    \end{subfigure} \setlength{\belowcaptionskip}{-6pt}
    \caption{Data set Characteristics. 
    Distribution of plans with plan length for home and factory domains. Frequency of interaction of top 10 objects and frequency of actions for top 10 actions. 
    The collected data set contains diverse interactions in complex spaces.
    }
    \label{fig:dataset}
\end{figure}

To curate an imitation learning dataset, we recruit human instructors and provide them with goals. They instruct the robot by specifying a sequence of symbolic actions (one at a time) to achieve each goal. Each action is simulated so that they can observe its effects and the new world state. We encourage the instructors to to complete the task as quickly as possible, making use of available tools in the environment. To familiarize them with the simulation platform, 
we conduct tutorial sessions before data collection. 
%
%
%
%
Our resulting dataset consists of diverse plans with different action sets and object interactions.
We collected plan traces from $\mathrm{12}$ human subjects using  
domain randomization with $\mathrm{10}$ scenes and $8$ semantic goals resulting in 
a corpus of $\mathrm{708}$ and $\mathrm{784}$ plans for the 
home and factory domains. 
%
%
%
%
Figure~\ref{fig:num_int} and \ref{fig:num_actions} show number of interactions with 10 most interacted objects and frequency of 10 most frequent actions respectively. 
The complete set of objects and goals is given in Table~\ref{tab:dataset_complete}.
We also perform data augmentation  by perturbing the metric states in the 
human plan trace, performing random replacements of scene objects and validating plan feasibility in simulation. 
The process results in $\mathrm{3540}$ and $\mathrm{3920}$ plans, respectively. 
%
%
%
Variation was observed in tool use for an instructor for different 
goals, and within similar goals based on context. 
%
%
%
The annotated corpus was split as $(75\%:25\%)$ forming the Training data set and the Test data set  
to evaluate model accuracy. 
A $10\%$ fraction of the training data was used as the 
Validation set for hyper-parameter search. 
No data augmentation was performed on the Test set. 


\subsection{Generalization Test Set}  
%
%
In order to asses the model's capacity to generalize to unseen worlds, we curate a second 
test environment populated by instances of novel object types 
placed at randomized locations. 
The following sampling strategies were used: 
(i) \emph{Position}: perturbing and exchanging object positions in a scene.
(ii) \emph{Alternate}: removal of the most frequently used tool in demonstrated plans
evaluating the ability to predict the alternative, next best tool to use.
(iii) \emph{Unseen}: replacing an object with a similar object, which is not present in training.
(iv) \emph{Random}: replacing a tool with a randomly picked object which is \emph{unrelated} to the task.
(v) \emph{Goal}: replacing the \blue{goal objects} (objects included as part of the goal specifications). For example: milk and fridge in the goal ``put milk inside fridge''. Here milk may be replaced with another similar object, such as apple and fridge by a container, such as cupboard.
This process resulted in a Generalization Test set with $7460$ (goal, plan) pairs.  
%

%% file: experiments.tex
\section{Experiments}
\label{sec:experiments}








We first present the results corresponding to high-level tool prediction that leverages an updated \textsc{ToolNet} model. We then present the goal-reaching performance of \textsc{Tango} and the unified \modelname model. Finally, we present a detailed analysis of the resulting plans in terms of generalization, robustness to execution errors and plan efficiency.

\subsection{Sequential Tool Prediction}

\subsubsection{Baseline}
We compare against the basic GGCN model of Section~\ref{sec:semantic} as our baseline model. This model is similar to the \textit{ResActGraph} model proposed by \citeA{liao2019synthesizing} and its encoder incorporates technical ideas from recent imitation learning works \cite{shridhar2019alfred} on action prediction. In the subsequent discussion, we consider similar size of the hyperparameter set across all models performing the same task.

\subsubsection{Evaluation Metrics}
We first compare the accuracy of the updated \textsc{ToolNet} model on the test and generalization test sets.  We test \textsc{ToolNet}'s tool prediction capabilities in two settings. In the first setting, we use the dataset as described in Section~\ref{sec:data} and split it according to the scene instance, \textit{i.e.}, home and factory. 
We use accuracy as our performance measure. Here, a tool prediction is deemed correct if the predicted tool is used in at least one of the various annotated plans for the $(\rm{goal,scene})$ pair and incorrect otherwise.

\begin{table*}[!t]
    \centering
    \resizebox{\textwidth}{!}{
    \begin{tabular}{@{}c !{\vrule width1pt}c|c !{\vrule width1pt}c|c !{\vrule width1pt}c|c|c|c|c@{}}
    \toprule 
    \multirow{2}{*}{\textbf{Model}} & \multicolumn{2}{c !{\vrule width0.8pt}}{\textbf{Test Set}} & \multicolumn{2}{c !{\vrule width0.8pt}}{\textbf{Generalization}} & \multicolumn{5}{c}{\textbf{Generalization Test cases}}\tabularnewline
    \cline{2-10} 
     & Home & Factory & Home & Factory & Position & Alternate & Unseen & Random & Goal\tabularnewline
    \midrule
    Baseline (GGCN) & 32.01 & 20.89 & 21.22 & 19.50 & 8.73 & 6.29 & 12.22 & 9.81 & 46.31\tabularnewline
    Updated \textsc{ToolNet} & \textbf{79.87} & \textbf{75.12} & \textbf{80.16} & \textbf{78.01} & \textbf{75.55} & \textbf{58.83} & \textbf{60.74} & \textbf{49.97} & \textbf{94.10}\tabularnewline
    \bottomrule
    \end{tabular}}
    \caption{A comparison of the updated \textsc{ToolNet} model with its previous version and GGCN baseline for the accuracy of prediction of tool sequences instead of single tool. \blue{Highest scores are shown in bold.}}
    \label{tab:toolnet_comparison}
\end{table*}

\subsubsection{Results}
Table~\ref{tab:toolnet_comparison} shows the final accuracies on the test-set  and generalization test-set 
with individual accuracies for each test-type. On the regular test set,  \textsc{ToolNet} outperforms the baseline by 47.87 \blue{(149\% higher)} and 54.23 \blue{(259\% higher)} accuracy points on Home and Factory domains, respectively. 

A similar pattern is found in the generalization test set, where each improvement is 58.94 \blue{(278\% higher)} and 58.51 \blue{(300\% higher)} accuracy points for Home and Factory domains. The reasoning behind the improvements is as follows. The goal-conditioned attention gives major improvement in the \textit{Goal} generalization test type, since goal objects get replaced in those examples. Explicitly biasing the model to use the features of those objects (through conditioned attention) increases their importance, and likely reduces overfitting. An example for \textit{Alternate} case is when \emph{generator} is specified in the goal, and wood (the fuel for generator, and the most likely tool) is made absent from the scene. The model could err in  giving attention to the \emph{wood-cutter} tool, which is often correlated with wood. However, conditioned attention gives low attention to \emph{wood-cutter} and predicts \emph{gasoline}, instead. 

Moreover, factored likelihood predictions helps the most in \textit{Unseen} cases, since without this component, the model cannot predict any unseen tool. A decent performance of earlier models in this case is attributed to alternative possible correct answers (any alternative seen tool or no-tool) due to multiple annotations per scenario.
The $\mathrm{ConceptNet}$ embeddings likely contain commonsense knowledge about unseen tools and objects, for example, whether a new tool is flat or not (which should help in ascertaining whether it can be used for transport or not). Using these embeddings makes huge improvement in \textit{Unseen} cases where entirely new objects are to be predicted as tools. 
Finally, giving higher weight to optimal plans allows the model to differentiate tools by plan execution time and not human usage frequency.
This helps in improved metric generalization, predicting nearby tools in the \textit{Position} test case. Overall, the complete architecture provides the maximum generalization accuracy among all models.

\subsection{Action Prediction}

\subsubsection{Baselines} 
We compare to the following three baseline models.  (1) \emph{ResActGraph} model ~\cite{liao2019synthesizing}, 
    augmented with $\mathrm{FastText}$ embeddings~\cite{mikolov2018advances}. (2)  \emph{Affordance-only} baseline inspired from ~\cite{hermans2011affordance} that learns a 
    co-association between tasks and tools, implemented by 
    excluding the graph convolutions and attention from \textsc{Tango}. (3) \emph{Vanilla Deep Q-Learning (DQN)} approach ~\cite{bae2019multi} that 
    learns purely by interactions with a simulator, receiving positive reward for reaching a goal state.

\subsubsection{Tool Prediction Accuracy}

Our experiments use the following accuracy metrics for model evaluation: 
(i) \emph{Action prediction accuracy}: the fraction of tool interactions predicted by the model that 
matched the human demonstrated action $a_t$ for a given state $s_t$, and 
(ii) \emph{Plan execution accuracy}: the fraction of estimated plans that are successful, i.e., can be executed by the robot in simulation and attain the intended goal (in max. $50$ steps). 

We first present the results for the vanilla \textsc{Tango} model and later present improvements with the \textsc{ToolTango}.

\subsubsection{Comparing \textsc{Tango} with Baselines}

Table~\ref{tab:ablation} (top half) compares the \textsc{Tango} model performance with the baseline models. 
The \textsc{Tango} model shows a $14-23$ point increase in \emph{Action prediction accuracy} 
and a $66-71$ points increase in the 
\emph{Plan execution accuracy} when compared to the \emph{ResActGraph} baseline.  
Note that the \textit{ResActGraph} model learns a scene representation assuming a fixed and known set of object types 
and hence can only generalize to new randomized scenes of known objects.  
%
%
In contrast, the \textsc{Tango} model can not only generalize to randomized scenes with known object types 
(sharing the GGCN backbone with \textit{ResActGraph}) but can to novel scenes new object types  
(relying on dense semantic embeddings) and an a-priori unknown number of instances (enabled by a factored likelihood). 

The \emph{Affordance-only} baseline model is confined to learning the possible association between a 
tool object type and the task specified by the human (largely ignoring the environment context). 
This approach addresses only a part of our problem as it ignores the sequential decision making aspect, where 
tools may need to be used in sequence to attain a goal. 
Finally, the vanilla DQN baseline achieves less than $20\%$ policy accuracy (even after a week of training). 
In contrast, the \textsc{Tango} model shows accurate results after training on imitation data for $12-18$ hours. 
The challenges in scaling can be attributed to the problem size ($\approx$1000 actions), long plans, sparse and delayed rewards 
(no reward until goal attainment).

\begin{table*}[!t]
    \centering
    \resizebox{\textwidth}{!}{
    \begin{tabular}{@{}c !{\vrule width1pt}c|c !{\vrule width1pt}c|c !{\vrule width1pt}c|c|c|c|c|c|c@{}}
    \toprule 
    \multirow{2}{*}{\textbf{Model}} & \multicolumn{2}{c !{\vrule width0.8pt}}{\textbf{Action Prediction}} & \multicolumn{2}{c !{\vrule width0.8pt}}{\textbf{Plan Execution}} & \multicolumn{7}{c}{\textbf{Generalization Plan Execution Accuracy}}\tabularnewline
    \cline{2-12} 
     & Home & Factory & Home & Factory & Home & Factory & Position & Alternate & Unseen & Random & Goal\tabularnewline
    \midrule
    Baseline (ResActGraph) & 27.67 & 45.81 & 26.15 & 0.00 & 12.38 & 0.00 & 0.00 & 0.00 & 0.00 & 25.10 & 9.12\tabularnewline
    Affordance Only & 46.22 & 52.71 & 52.12 & 20.39 & 44.10 & 4.82 & 17.84 & 47.33 & 29.31 & 29.57 & 34.85\tabularnewline
    DQN  & - & - & 24.82 & 17.77 & 15.26 & 2.23 & 0.00 & 0.00 & 12.75 & 9.67 & 4.21\tabularnewline
    \textsc{Tango} & 59.43 & 60.22 & 92.31 & 71.42 & \textbf{91.30} & \textbf{60.49} & \textbf{93.44} & \textbf{77.47} & \textbf{81.60} & 59.68 & \textbf{59.41}\tabularnewline
    \midrule
    \multicolumn{12}{c}{\textbf{Model Ablations}}\tabularnewline
    \midrule
    - GGCN (World Representation) & 59.43 & 60.59 & 84.61 & 27.27 & 78.02 & 38.70 & 70.42 & 58.79 & 60.00 & 56.35 & 38.64\tabularnewline
    - Metric (World Representation) & 58.8 & 60.84 & 84.61 & 62.34 & 72.42 & 51.83 & 59.68 & 67.19 & 60.79 & \textbf{84.47} & 21.70\tabularnewline
    - Goal-Conditioned Attn & 53.14 & 60.35 & 53.85 & 11.69 & 37.02 & 8.80 & 35.33 & 15.05 & 32.14 & 41.67 & 6.51\tabularnewline
    - Temporal Action History & 45.91 & 49.94 & 24.61 & 0.00 & 8.55 & 0.00 & 0.00 & 0.00 & 0.00 & 30.56 & 1.15\tabularnewline
    - Factored Likelihood & 61.32 & \textbf{61.34} & \textbf{95.38} & \textbf{85.71} & 34.22 & 43.44 & 90.50 & 14.82 & 30.65 & 64.67 & 53.26\tabularnewline
    - ConceptNet & \textbf{63.52} & 60.35 & 89.23 & 57.14 & 81.86 & 56.97 & 82.33 & 68.61 & 74.57 & 65.73 & 47.92\tabularnewline
    - Constraints & 57.23 & 57.74 & 64.62 & 37.66 & 62.98 & 41.95 & 84.95 & 45.39 & 39.99 & 36.11 & 84.85\tabularnewline
    - Auto-regression & 56.60 & 60.22 & 69.23 & 50.65 & 73.75 & 53.32 & 71.24 & 66.43 & 61.27 & 70.51 & 34.66\tabularnewline \bottomrule 
    \end{tabular}}
    \caption{{A comparison of \emph{Action prediction} and \emph{Plan execution} accuracies for the baseline, the proposed \textsc{Tango} model, and ablations. Results are presented for test and generalization data sets (under five sampling strategies) derived from the home and factory domains. \textit{Accuracy Prediction} is the percentage of predicted actions matching the human input on the Test set. \textit{Plan Execution} is the percentage of plans successfully executed in the Test set. \textit{Generalization Plan Accuracy} is the percentage of plans successfully executed in the  set. \blue{Highest scores shown in bold.}
    }}
    \label{tab:ablation}
\end{table*}

Next, we assess the zero-shot transfer setting, i.e., 
whether the model can perform common sense generalization in worlds with new objects unseen in training.
The same table shows that the plans predicted 
by \textsc{Tango} lead to an increase of up to $56$ points in plan execution accuracy on Generalization Test set 
over the best-performing baseline model.
This demonstrates accurate prediction and use of unseen tool objects for a given goal. 
%
%
Specifically, in the home domain, 
if the \emph{stool} is not present in the scene, 
the model is able to use a \emph{stick} instead to fetch far-away objects. 
Similarly, if the robot can predict the use of a box for transporting objects even if it 
has only seen the use of a tray for moving objects during training. 
The \textit{ResActGraph} model is unable to adapt to novel worlds and obtains zero points in several generalization tests. 

The poorer performance of the \emph{Affordance-only} model can again be attributed to the fact that 
planning tool interactions involves sequential decision-making. 
Even if the robot can use affordance similarity 
to replace a \emph{tray} object with a \emph{box}, it still needs to predict 
the opening of the \emph{box} before placing an item in its plan for a successful execution. 
This is corroborated by the drop in performance for the \textit{Unseen} generalization tests
for this model by $52.3$ points. 
Finally, the vanilla DQN model lacks a clear mechanism for transferring to novel settings, hence 
shows poor generalization in our experiments. 
\subsubsection{Ablation Analysis of \textsc{Tango} Components} 
We analyze the importance of each component of the \textsc{Tango} model 
by performing an ablation study. 
Table~\ref{tab:ablation} (lower half) presents the results.
For a fair comparison, the model capacities remain the same during the ablation experiments. 

The model builds on the \emph{GGCN} environment representation 
encoding the inter-object and agent-object relational properties. 
The ablation of the GGCN component results in a 
reduction of 22\% in the generalization accuracy in the factory domain 
(where tools may be placed at multiple levels in the factory). 
The inclusion of this component allows the robot to leverage 
relational properties such as OnTop to predict the use of 
tools such as a ramp to negotiate an elevation or a stick to 
fetch an object immediately beyond the manipulator's reach. 

The \emph{Metric} component encodes the metric properties of objects 
in the scene such as positions, size etc. 
Experiments demonstrate its effectiveness in prediction tool interactions based on relative 
sizes of interacting objects. 
E.g., the model successfully predicts that \emph{fruits} can be transported  
using a \emph{tray} but larger \emph{cartons} require a \emph{box} for the same task.
The ablation of this component leads to a reduction of $10.2$ points in the 
\textit{Alternate} generalization tests 
as the ablated model unable to adapt the tool when there are unseen 
objects with different sizes than those seen during training. 

Next, we assess the impact of removing the \emph{Goal-Conditioned Attention component}. 
This experiment shows a a significant reduction ($\approx 50$ points) in 
the \emph{Plan execution accuracy} on the Generalization Test set, 
particularly in scenes with a large number of objects. 
The attention mechanism allows learning of a restricted context of tool objects 
that may be useful for attaining the provided goal, 
in essence, filtering away goal-irrelevant objects populating the scene. 
Additionally, note that the inclusion of this component allows tool predictions 
to be \emph{goal-aware}.
Consequently, we observe ablating this component leads to a reduction of 
$53$ points in the \textit{Goal} generalization test set where 
the goal objects are perturbed.  

%
The \emph{Action History} component utilizes the agent's past interactions 
for the purpose of predicting the next tool interaction. 
The inclusion of this component allows learning of 
correlated and commonly repeated action sequences. 
For instance, the task of exiting from a room typically involves a plan fragment 
that includes moving to a door, opening it and exiting from the door and 
are commonly observed in a number of longer plans. 
The ablation of this component leads to erroneous predictions 
where a particular action in a common plan fragment is missing or incorrectly predicted. 
E.g., a robot attempting to pick an object inside an enclosure without opening the lid. 
In our experiments, we observe that ablating the model 
leads to a significant decrease in goal reach-ability, causing a $70$ point decrease in the 
\emph{Plan Execution accuracy} and $72$ point drop in the \emph{Generalization accuracy}. 

The need for generalization to novel scenes implies that our model cannot assume  
a fixed and a-priori known set of objects that the robot can interact with. 
Generalization to an arbitrary number of objects in the scene is accomplished  
by factoring model predictions over individual objects in a recurrent manner. 
Ablating the factored likelihood components results in a simpler model that performs predictions 
over a known fixed-size object set. 
The simplified model displays a higher action-prediction and plan-execution accuracies in 
known world. 
Crucially, we observe that ablating this component results in a significant decrease of   
$51$ and $63$  points in the \textit{Unseen} and the \textit{Alternate} generalization test sets. 
%
 
Finally, the \emph{ConceptNet embeddings} are important 
for semantic generalization to unseen tool types. 
%
%
We replace ConceptNet embeddings with $\mathrm{FastText}$~\cite{mikolov2018advances} embeddings in the \emph{-ConceptNet} model to show their importance.
The \emph{-ConceptNet} model shows poorer generalization (6.5\% decrease) as it models
word affinity as expressed in language only. 
$\mathrm{ConceptNet}$ embedding space models relational affinity between objects as present in the knowledge-base.

\begin{table*}[!t]
    \centering
    \resizebox{\textwidth}{!}{
    \begin{tabular}{@{}c !{\vrule width1pt}c|c !{\vrule width1pt}c|c !{\vrule width1pt}c|c|c|c|c|c|c@{}}
    \toprule 
    \multirow{2}{*}{\textbf{Model}} & \multicolumn{2}{c !{\vrule width0.8pt}}{\textbf{Action Prediction}} & \multicolumn{2}{c !{\vrule width0.8pt}}{\textbf{Plan Execution}} & \multicolumn{7}{c}{\textbf{Generalization Plan Execution Accuracy}}\tabularnewline
    \cline{2-12} 
     & Home & Factory & Home & Factory & Home & Factory & Position & Alternate & Unseen & Random & Goal\tabularnewline
    \midrule
    \textsc{TANGO} & 59.43 & 60.22 & 92.31 & 71.42 & 91.30 & 60.49 & 93.44 & 77.47 & 81.60 & 59.68 & 59.41\tabularnewline
    \textsc{ToolTANGO} & \textbf{64.12} & \textbf{63.72} & \textbf{95.69} & \textbf{77.01} & \textbf{92.88} & \textbf{62.97} & \textbf{94.38} & \textbf{88.12} & \textbf{92.71} & \textbf{78.01} & \textbf{64.43}\tabularnewline
    \bottomrule
    \end{tabular}}
    \caption{A comparison of \textsc{TANGO} with \textsc{ToolTANGO} model.}
    \label{tab:tooltango}
\end{table*}

\subsubsection{Comparing \textsc{Tango} with \modelname}

Table~\ref{tab:tooltango} compares the performance of \modelname with the \textsc{Tango} model. \modelname gives improved scores in both domains and for every generalization test case. The \textit{Action prediction} accuracy of \modelname is higher than \textsc{Tango} by 4.69 and 3.50 points for Home and Factory domains, respectively. Similarly, we see an increase in the \textit{Plan execution accuracy} both for Test and Generalization Test sets. We get 3.38- 5.59 higher points on Test and 1.58-2.48 points in Generalization Test set for Home and Factory domains. The improvement is higher in Factory domain where the probability of using multiple tools in the same plan to reach a goal state is higher. This is particularly due to the independent tool prediction in \modelname that aids action prediction in complex settings requiring longer action sequences (more details in Section~\ref{sec:analysis}). The performance improvement is highest for the \textit{Alternate} and \textit{Random} cases where the tools are replaced by an alternate tool or another non-tool object. This shows the importance of independent tool-likelihood score prediction in the \modelname model.

\begin{figure}[!t]
    \centering
    \begin{minipage}{\columnwidth}
        \centering
        \setlength{\belowcaptionskip}{-5pt}
        \includegraphics[width=\textwidth]{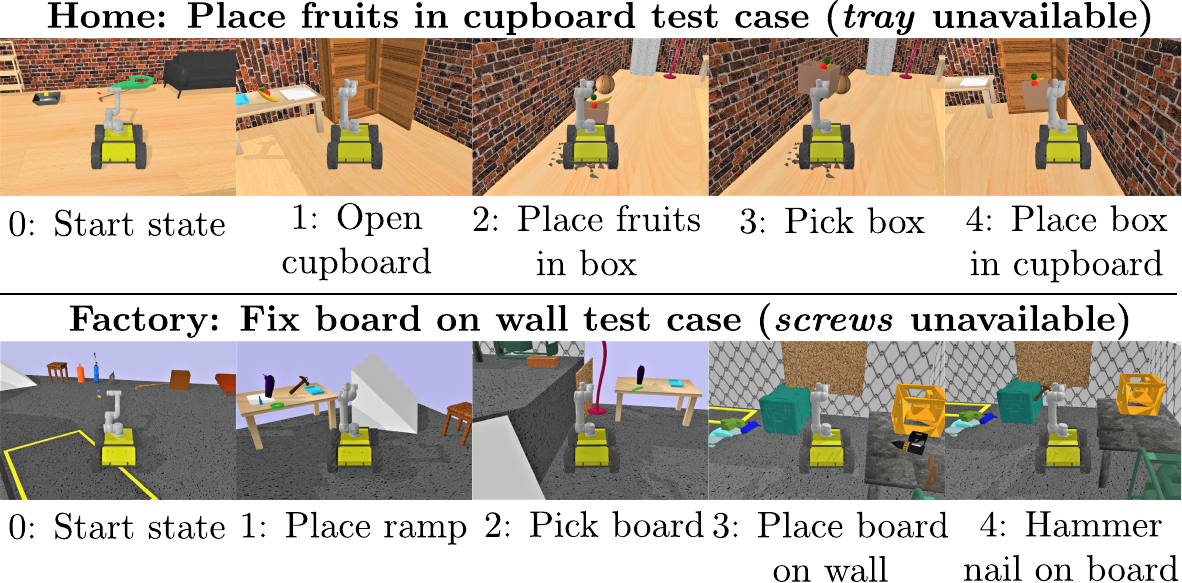}
        \caption{
        {
        A simulated robot manipulator uses \modelname to synthesize tool interactions in 
        novel contexts with unseen objects. (top) \modelname predicts the use of a \textit{box} when \textit{tray} is unavailable. (bottom) \modelname predicts the use of \textit{hammer} and \textit{nails} when \textit{screws} are unavailable. 
        }}
        \label{fig:plans}
    \end{minipage}\ 
\end{figure}

\begin{figure}[!t]
        \centering
        \setlength{\belowcaptionskip}{-8pt}
        \includegraphics[width=0.7\linewidth]{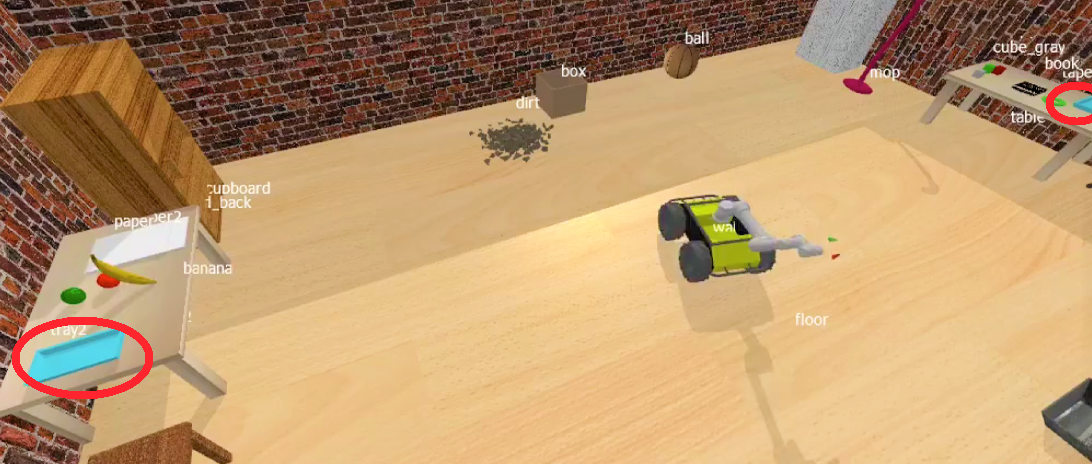}
        \caption{{The model predicts the instance of \textit{tray} (on the left) which is closer to the \textit{fruits} (goal objects) other instance (one on the right). }} 
        \label{fig:metric_prop}
\end{figure}
\begin{figure}
        \centering
        \setlength{\belowcaptionskip}{-8pt}
        \includegraphics[width=0.7\linewidth]{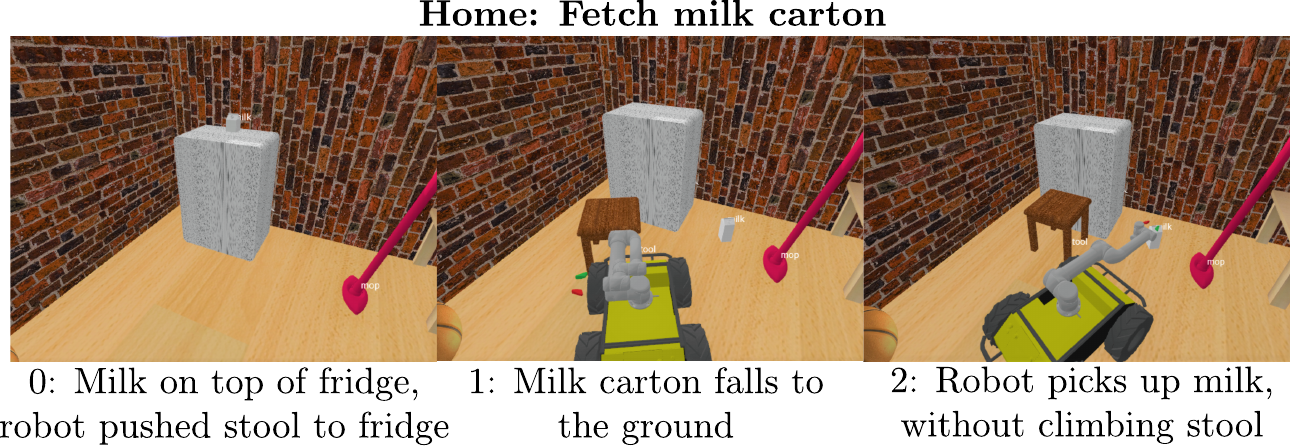}
        \includegraphics[width=0.7\textwidth]{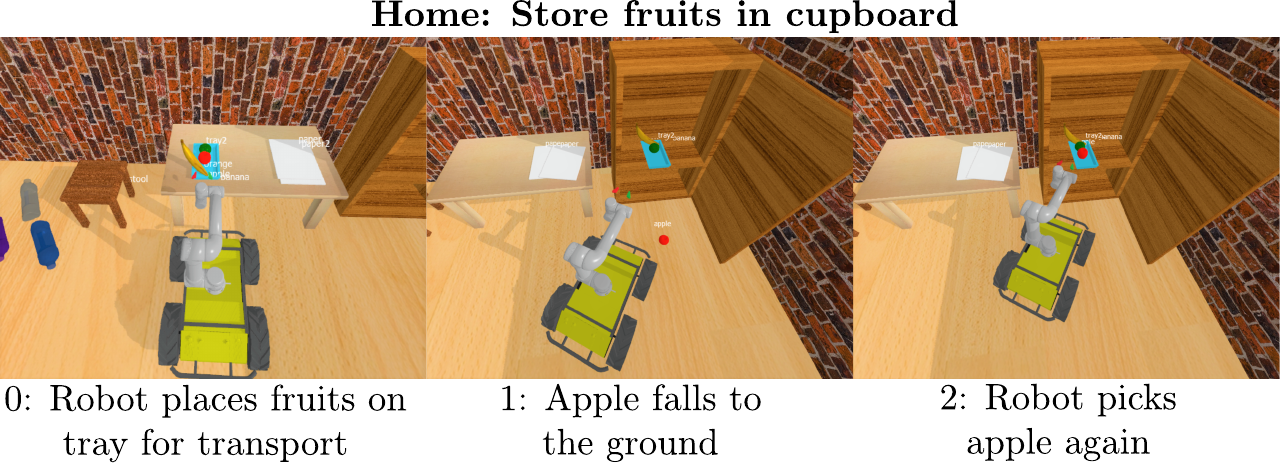}
        \caption{{Interleaved action prediction and 
        execution enables adaptation in case of unexpected errors during action execution. (top) robot recovers after the milk carton falls. (bottom) recovers after a fruit falls from the \textit{tray} by picking it up again.}} 
        \label{fig:robust}
\end{figure}

\subsection{Analysis of Resulting Plans}
\label{sec:analysis}

\subsubsection{Evidence of Generalization}
Figure \ref{fig:plans} shows the robot using the learned model to 
synthesize a plan for a declarative goal. 
Here, if the goal is to transport fruits and human demonstrates usage of \emph{tray} and the model never sees \emph{box} while training, \modelname uses \emph{box} in a scene where tray is absent, showing that it is able to predict semantically similar tools for task completion. 
Similarly, for the goal of fixing a board on the wall, if humans use \emph{screws} the agent uses \emph{nails} and \emph{hammer} when screws are absent from the scene. 
Figure~\ref{fig:metric_prop} shows how the model 
uses the position information of tool objects to predict the 
tool closer to the goal object or the agent. 
The world representation encodes the metric properties of objects 
(position and orientation) that allows the robot to 
interact with nearer tool objects. 

\begin{figure}
        \centering \setlength{\belowcaptionskip}{-8pt}
        \includegraphics[width=0.55\linewidth]{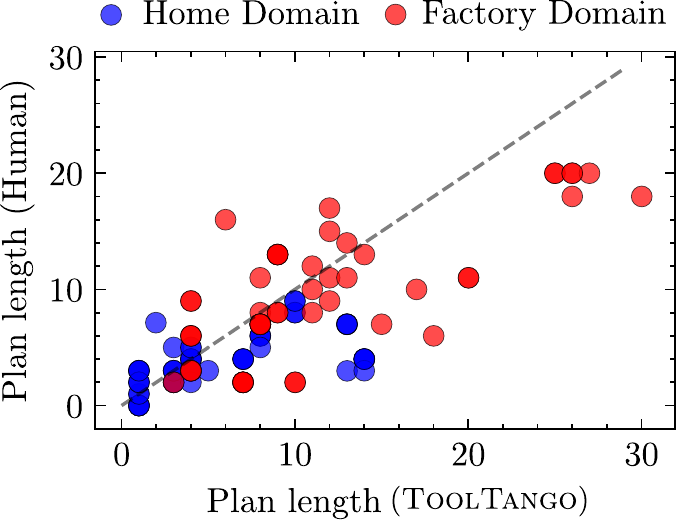}
        \caption{{Scatter plot comparing the lengths of plans obtained from model predictions and those from human demonstrations. The plans generated from \modelname are of similar length as that of human demonstrators.}} 
        \label{fig:planlen2}
\end{figure}

\subsubsection{Robustness to Errors}
Figure~\ref{fig:robust} shows the robustness to unexpected errors and stochasticity in action execution.  
Consider the task of ``fetching a carton", where the milk carton is on an elevated platform, 
the model predicts the uses a stool to elevate itself. 
The carton falls due to errors during action execution. 
Following which, the robot infers that the stool is no longer required and directly fetches the carton. 
Similarly, for the task of ``storing away the fruits in the cupboard", the robot predicts the use of tray for the transport task. 
During execution the apple falls off the tray. The robot correctly re-collects the apple.   


\subsubsection{Plan Efficiency Comparison}

Figure~\ref{fig:planlen2} compares the length of robot plans 
predicted by the learned model against human demonstrated plans.
We observe that, on average, the predicted plan lengths are 
close to the human demonstrated ones. 
In 12\% cases, the plans predicted by \modelname utilize 
tools satisfying the goal condition 
in fewer steps compared to the human demonstrated plan. 
%
%
%
%
%
\begin{figure}[!t]
        \centering 
        \includegraphics[width=0.9\linewidth]{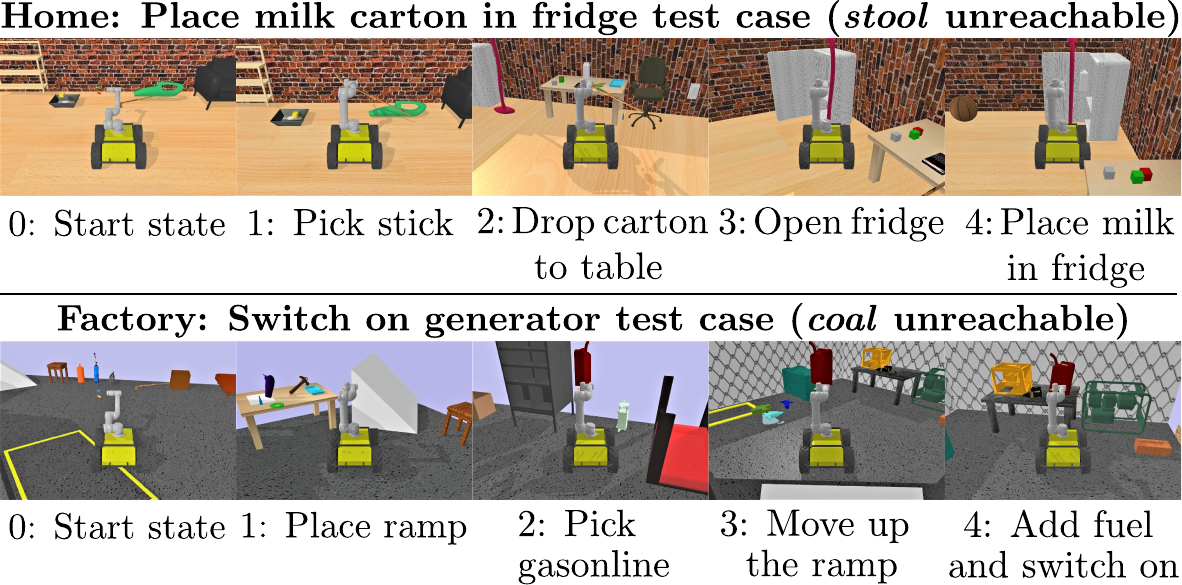}
        \captionof{figure}{Test cases where \textsc{TANGO} fails but \textsc{ToolTANGO} reaches the goal state. Typically in cases with complex and long plans with multiple tools being used. \textsc{Tango} predicts \textit{stool} in the first case and \textit{coal} in the second case, both of which are unreachable. \modelname predicts reachable tools.
        (Top) Here the robot needs to place the milk carton which is at a higher location which is unreachable without a stool. In this test case, the stool in unreachable so the model uses a stick to reach the milk carton. (Bottom) Switching on the generator requires a fuel. In this test case coal is unreachable/unavailable and so the model instead uses gasoline to turn on the generator.}
        \label{fig:correction}
\end{figure}
\begin{figure}[!t]
        \centering
        \includegraphics[width=0.62\linewidth]{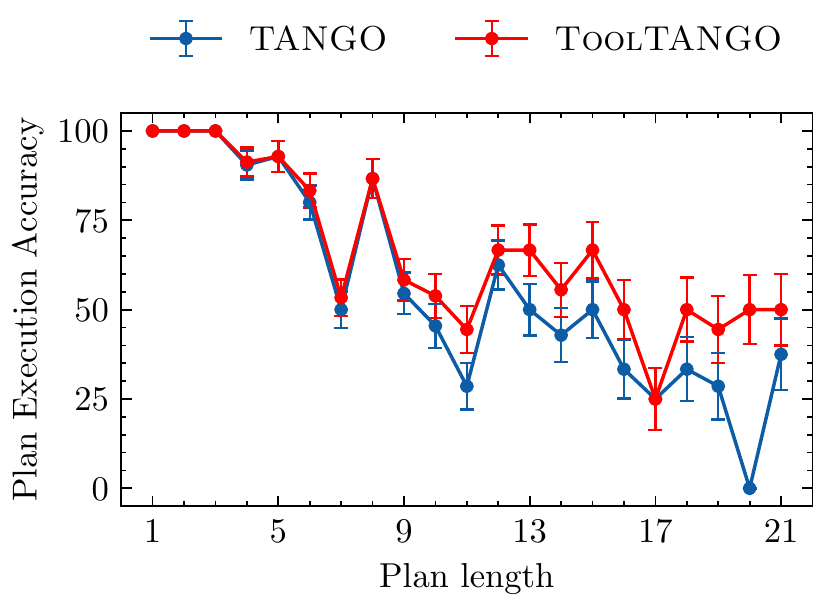}
        \captionof{figure}{{Execution accuracy of inferred plans with plan length for \textsc{Tango} and \modelname. The latter gives a higher goal-reaching performance than the former for plans with longer action sequences.}}
        \label{fig:test}
\end{figure}
\begin{figure}[!t]
        \centering
        \includegraphics[width=1.0\linewidth]{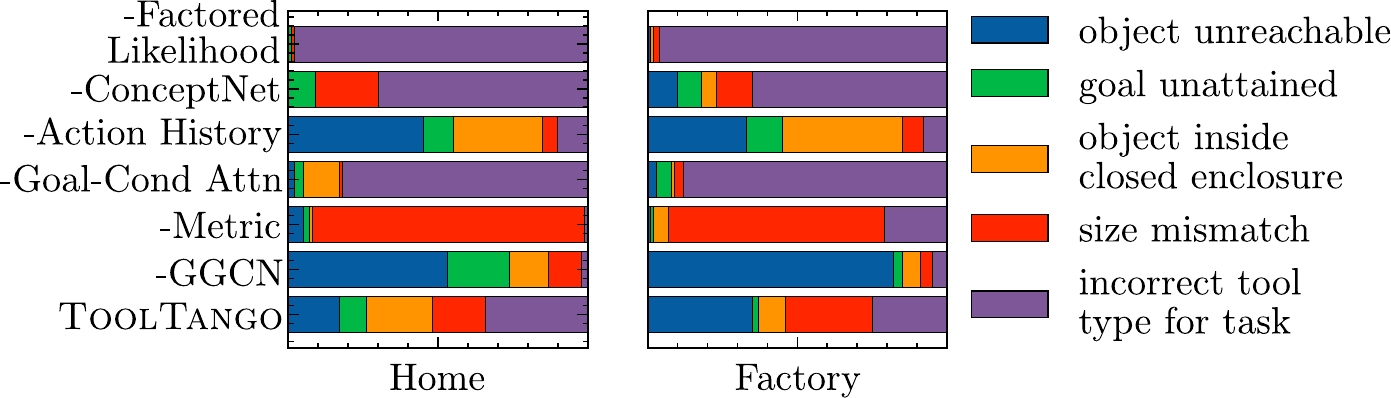}
      \captionof{figure}{{An analysis of fractional errors during plan execution using the learned \modelname model. \blue{The horizontal axis denotes the total errors for the ablated model. The absolute value of the total errors are shown in Table~\ref{tab:ablation}}}}
        \label{fig:errors}
\end{figure}
 
\subsubsection{Analysis of Independent Tool Prediction} 
Figure~\ref{fig:correction} demonstrates the advantage of independent tool prediction and how it augments the \modelname to improve goal-reaching performance.
The figure shows two cases where the \textsc{Tango} model is unable to reach the goal state, but \modelname reaches a goal. The first example shows a setting where a milk carton is placed on top of the fridge and stool is kept directly in front of the fridge. In this case, \textsc{Tango} predicts the use of \textit{stool} to reach the carton; however, it first opens the fridge door, making \textit{stool} inaccessible. On the other hand, \modelname predicts using a \textit{stick} to drop the carton on top of the table and is able to place it inside the fridge. The second example shows a case where the generator needs to be powered on after adding a fuel source. Here, \modelname predicts the picking \textit{coal} that is on top of the shelf. However, it does not use any tool to elevate itself and the execution returns an error of unreachable object. Instead, \modelname predicts the use of \textit{gasoline} that is placed on the ground and is successfully able to execute the plan.
 
Independent tool prediction also allows the model to perform better in complex settings requiring longer action sequences and using multiple tools.  Figure~\ref{fig:test} assesses the model accuracy 
with the lengths of the inferred plans. We observe that \modelname has a higher plan execution accuracy in case of (goal, scene) pairs where the average plan length to attain a goal state is high. For instance, if the goal is to place fruits in side the cupboard and all the fruits are on top of the fridge, the robot needs to use a tool to first reach the fruits and then use another tool to carry them to the cupboard. In this case the \textsc{Tango} model directly tries to pick unreachable fruits or predicts an incorrect tool to reach the fruits. \modelname is able to execute complex task by subsequently using a \textit{stool} and \textit{tray}.
 
%
\section{Limitations and Future Work}
\label{sec:limitations}

\subsection{Scaling to Longer Plans}
As we observed in Figure~\ref{fig:test}, the plan execution accuracy decreases by 20\% on the Test sets and 
30\% on Generalization Test sets. 
This merits investigation into planning abstractions \cite{vega2020asymptotically} 
for scaling to longer plan lengths in realistic domains.   
Figure~\ref{fig:errors} analyzes the errors encountered during plan execution using 
actions predicted by the proposed model. 
In 27\% of the cases, the model misses a pre-requisite actions 
required for a pre-condition for initiating the subsequent action. 
For example, missing the need to open the door 
before exiting the room (object unreachable $19\%$) or missing opening the cupboard before picking an object inside it (object inside enclosure $8\%$). 
There is scope for improvement here by incorporating explicit causal structure ~\cite{nair2019causal}. 

\subsection{Partially observable environments} 
\begin{figure}
    \centering \setlength{\belowcaptionskip}{-10pt}
    \includegraphics[width=0.6\columnwidth]{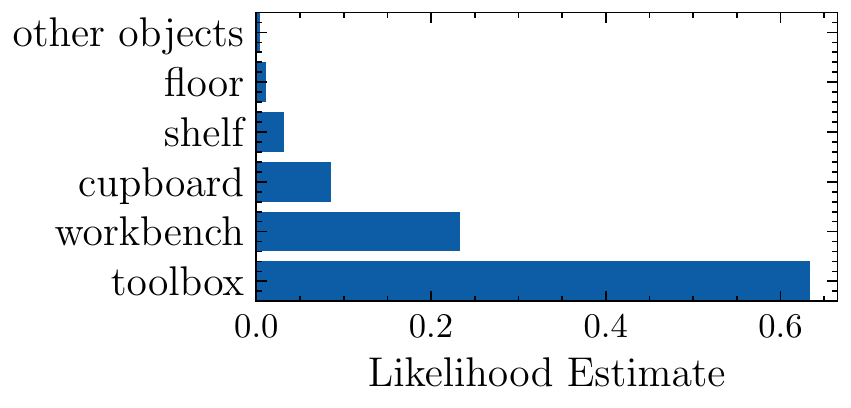}
    \caption{Likelihood estimates of exploration for different objects where ``screw'' might be found.}
    \label{fig:exploration}
\end{figure}

In realistic scenarios, the robot’s environment may only be partially-known
due to the limited view and scope of the robot’s sensors. For example,
tools such as screws may be stored away in a tool box and may not be
easily observed by the robot. In such a scenario, we expect the robot
to learn to predict possible  locations for exploration based on
common sense knowledge. For example, the robot should explore the tool
box. \blue{Further, the subgeometry information may be fed into the model to determine the set of objects in the environment.}
In order to address partially-observed worlds, we can extend the
prediction model as follows. Instead of learning attention only over
candidate objects, we can learn a \emph{generalized} attention over
spatial relations modeled in the graph network. Such an extension
allows the model to predict a preference order for locations that the
robot should explore to find the required tool.
Figure~\ref{fig:exploration} illustrates an example, where the 
robot
shows an example, where the robot predicts that the screws may
be found in toolbox or workbench. Please note that this result is indicative of
the possibility of using the model in partially-known world and will be
investigated in detail as part of future work.

\subsection{Extension to realistic robot control settings}

\blue{The action representations adopted in this work are inspired from task and motion planning communities~\cite{zhu2021hierarchical,migimatsu2020object}. We simulate a mobile manipulator (Clearpath Robotics Husky with a mounted Universal Robotics UR5 arm). The robot's action space is modeled as high-level skills (or behaviors), each in-turn realized using a low-level motion plan or controller that is parameterized at run time. For navigation actions, a standard state-space planner (A* with Manhattan distance as a heuristic) for a coarse collision free path and realize point to point motion through a velocity-based controller. For the manipulation, we consider a 6 degree of freedom model of the manipulator. For object manipulation a crane-grasp was implemented where the arm would move to a pre-grasp position to the intended object of interest. The arm motion was realized using a joint state controller moving the arm to the end-effector pose. The precise manipulation of tools is abstracted as follows: once the robot end-effector makes contact a rigid joint is established for subsequent motions till it is released. As mentioned in Section~\ref{sec:introduction}, learning the precise manipulation of tools is not the focus of this work and can be delegated to a method such as the one proposed by~\citeA{parkinferring}. All robot actions are executed in the PyBullet physics engine with a mesh-based model of objects and a URDF-model for the robot manipulation system. The technical details appear in Appendix~\ref{sec:manipulation}. The entire implementation of the robot behaviors and the simulation environment is available at~\url{https://github.com/reail-iitd/tango} for the use of the research community. }

%% file: appendix.tex
\section{Supplementary Material}
\label{appendix:supp_material}

All our code is available as a GitHub repository under BSD-2 License~\url{https://github.com/reail-iitd/tango}. Instructions for reproducing the results are given at~\url{https://github.com/reail-iitd/tango/wiki}.  A supplementary video demonstrating our data collection platform and model's generalization capability is available at~\url{https://www.youtube.com/watch?v=lUWU3rK1Gno}.

\section{Hyperparameter Details}
\label{appendix:hyperparams}
We detail the hyper-parameters for the \modelname architecture introduced 
in this paper. 

\begin{itemize}
\item 
\emph{Graph Structured World Representation. }
The Gated Graph Convolution Network (GGCN) was implemented with $4$-hidden layers, each of size $128$, with convolutions across $2$ time steps for every relation passing through a layer normalized GRU cell. 
The Parameterized ReLU activation function with a $0.25$ negative input slope was used in all hidden layers. 

\item 
\emph{Word Embeddings. }
The word embeddings (derived from $\mathrm{ConceptNet}$) were of size $300$. 
Additionally, the semantic state of each object was encoded as a one-hot vector of size $29$. Typically, there were 
$35$ and $45$ objects in the home and factory domains respectively. 

\item 
\emph{Fusing Metric Information. }
The metric encodings were generated from the metric information associated with objects using a $2$-layer Fully Connected Network (FCN) with $128$-sized layers. 

\item 
\emph{Encoding Action History. } 
A Long Short Term Memory (LSTM) layer of size $128$ was used to encode the action history using the generalized action encoding $\mathcal{A}(I_t(o^1_t, o^2_t))$. 

\item 
\emph{Goal-conditioned Attention. }
The attention network was realized as a $1$-layer FCN of layer size $128$ with a $\mathrm{softmax}$ layer at the end. 

\item 
\emph{Tool Prediction. }
To predict the tool-likelihood score $p_t$, a $3$-layer FCN was used, each hidden layer with size $128$ and output layer with size $1$ with sigmoid activation function. 

\item 
\emph{Action Prediction. }
To predict the action $I_t$, a $3$-layer FCN was used, each hidden layer with size $128$ and output layer with size $|\mathcal{I}|$. 
$I_t$ was converted to a one-hot encoding $\vec{I}_t$. This, with the object embedding $e_o$ was passed to the $o^1_t$ predictor via an FCN. This FCN consists of 3-hidden layers of size $128$ and a final layer of size $1$ with a sigmoid activation (for likelihood). 
The $\vec{I}_t$ and $o^1_t$ likelihoods were sent to the $o^2_t$ predictor to predict likelihoods for all object embeddings 
$e_o$. This part was realized as a $3$-layer FCN with hidden layer size $128$ and final layer of size $1$ with a sigmoid activation function. 

\item 
\emph{Training parameters.} 
Model training used a learning rate of $5\times 10^{-4}$. 
The Adam optimizer~\cite{kingma2014adam} with a weight decay parameter of $10^{-5}$ and a batch size of $1$ was used. 
An early stopping criterion was applied for convergence. 
The \emph{action prediction accuracy} was used as the comparison metric on the validation set or up to a maximum of $200$ epochs. 
\end{itemize}

\section{World Scenes}
\label{appendix:world_scenes}

\begin{figure}
    \centering
    \begin{subfigure}{0.49\linewidth}
        \centering
        \includegraphics[width=\linewidth]{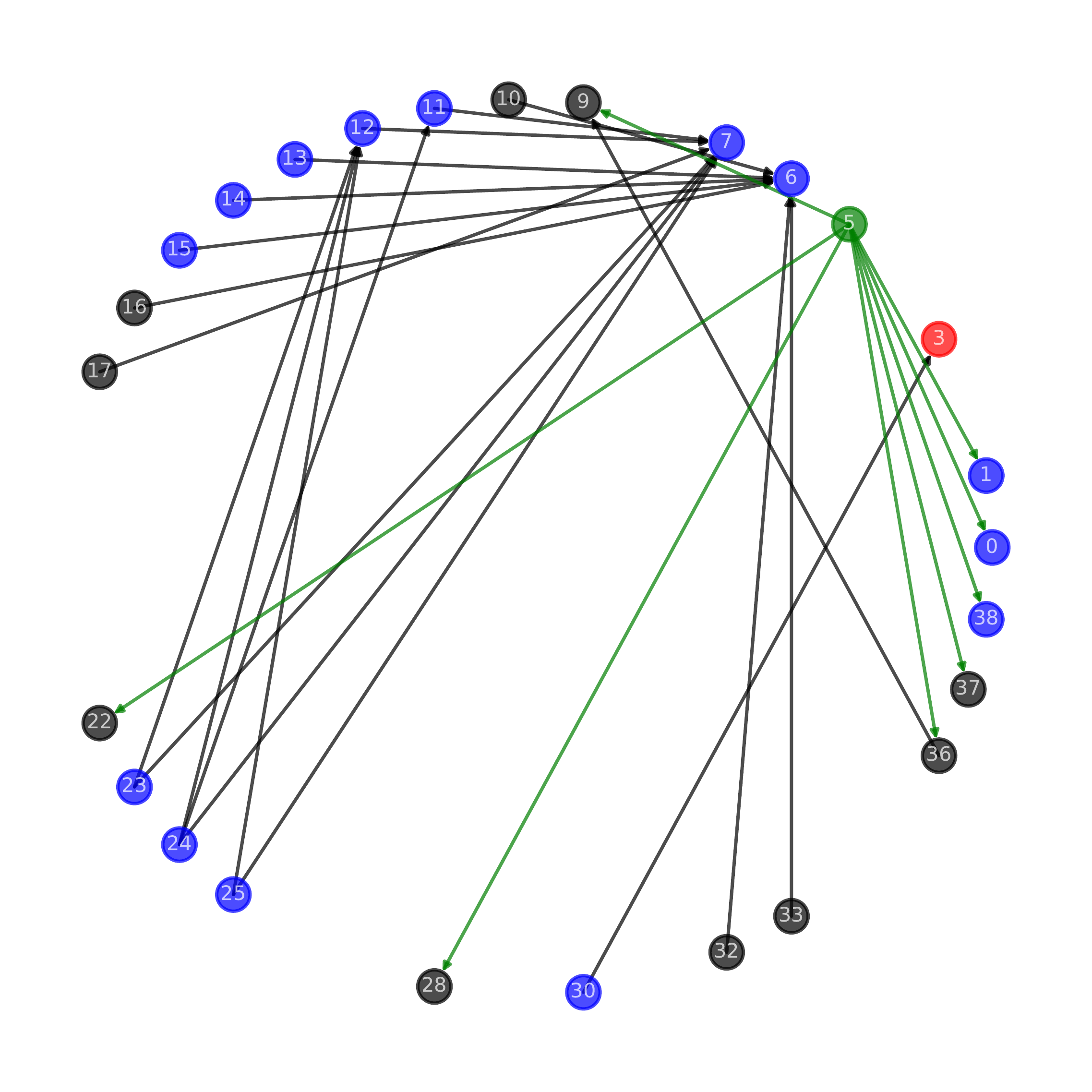}
        \caption{A sample home scene.}
    \end{subfigure}
    \begin{subfigure}{0.49\linewidth}
        \centering
        \includegraphics[width=\linewidth]{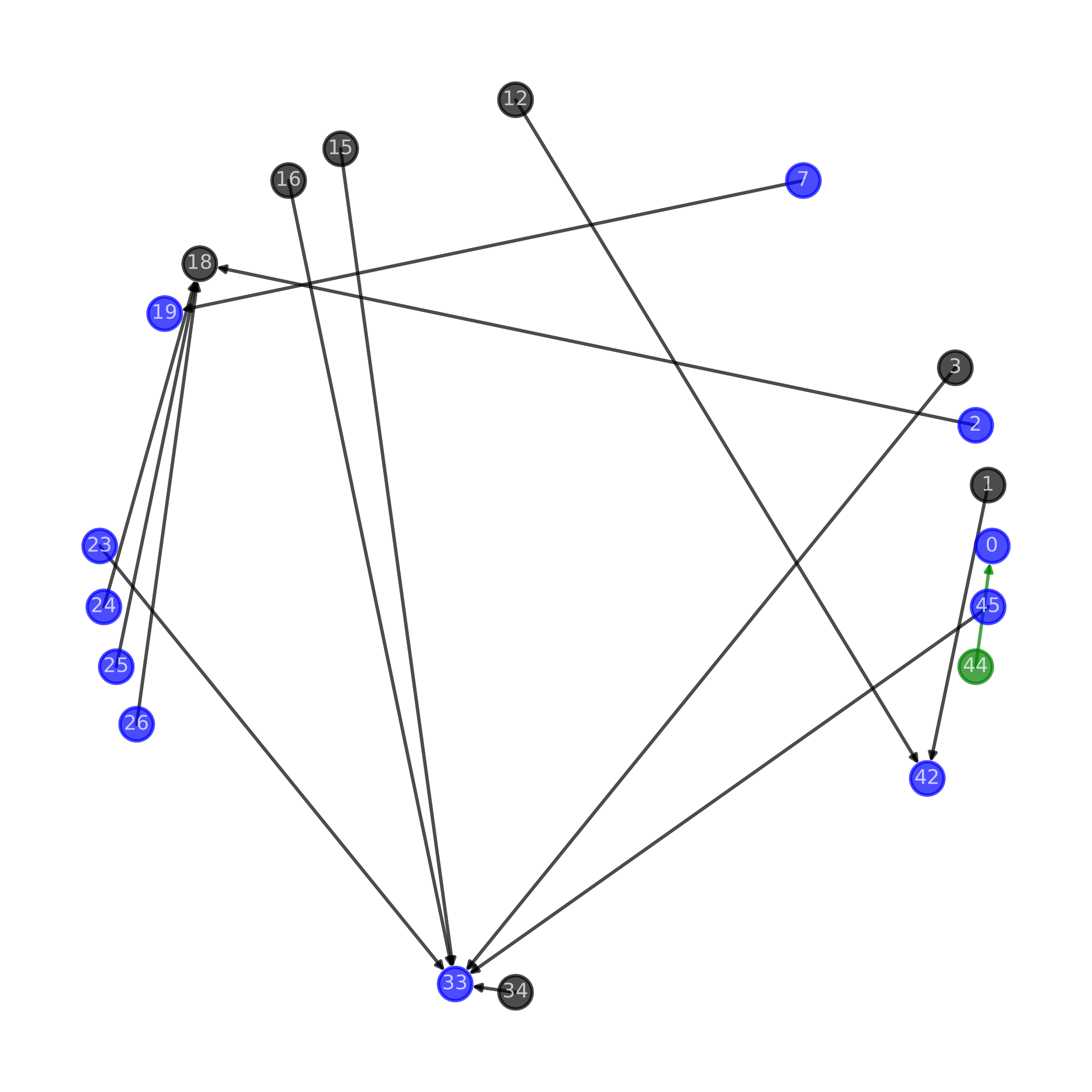}
        \caption{A sample factory scene.}
    \end{subfigure}
    \caption{\footnotesize \textbf{Sample World Scenes in Home and Factory domains}}
    \label{fig:world_home_scenes}
\end{figure}

The figure~\ref{fig:world_home_scenes} illustrates the object-centric graph representation of the 10 world scenes we used for Home and Factory domains. The \textit{agent node} is shown in green, \textit{tools} in black and \textit{objects with states} in red. The relations of \textit{Close} are shown in green (only populated for agent), \textit{On} in black, \textit{Inside} in red and \textit{Stuck to} in blue.

\noindent
The legend for node IDs to objects are given below:

\textbf{Home:} 0: floor, 1: walls, 2: door, 3: fridge, 4: cupboard, 5: husky, 6: table, 7: table2, 8: couch, 9: big-tray, 10: book, 11: paper, 12: paper2, 13: cube gray, 14: cube green, 15: cube red, 16: tray, 17: tray2, 18: light, 19: bottle blue, 20: bottle gray, 21: bottle red, 22: box, 23: apple, 24: orange, 25: banana, 26: chair, 27: ball, 28: stick, 29: dumpster, 30: milk, 31: shelf, 32: glue, 33: tape, 34: stool, 35: mop, 36: sponge, 37: vacuum, 38: dirt.

\textbf{Factory:} 0: floor warehouse, 1: 3D printer, 2: assembly station, 3: blow dryer, 4: board, 5: box, 6: brick, 7: coal, 8: crate green, 9: crate peach, 10: crate red, 11: cupboard, 12: drill, 13: gasoline, 14: generator, 15: glue, 16: hammer, 17: ladder, 18: lift, 19: long shelf, 20: mop, 21: nail, 22: oil, 23: paper, 24: part1, 25: part2, 26: part3, 27: platform, 28: screw, 29: screwdriver, 30: spraypaint, 31: stick, 32: stool, 33: table, 34: tape, 35: toolbox, 36: trolley, 37: wall warehouse, 38: water, 39: welder, 40: wood, 41: wood cutter, 42: worktable, 43: ramp, 44: husky, 45: tray.

\section{Realization of Object Relations}
\label{sec:manipulation}
\begin{figure}[!t]
    \centering
    \includegraphics[width=0.6\textwidth]{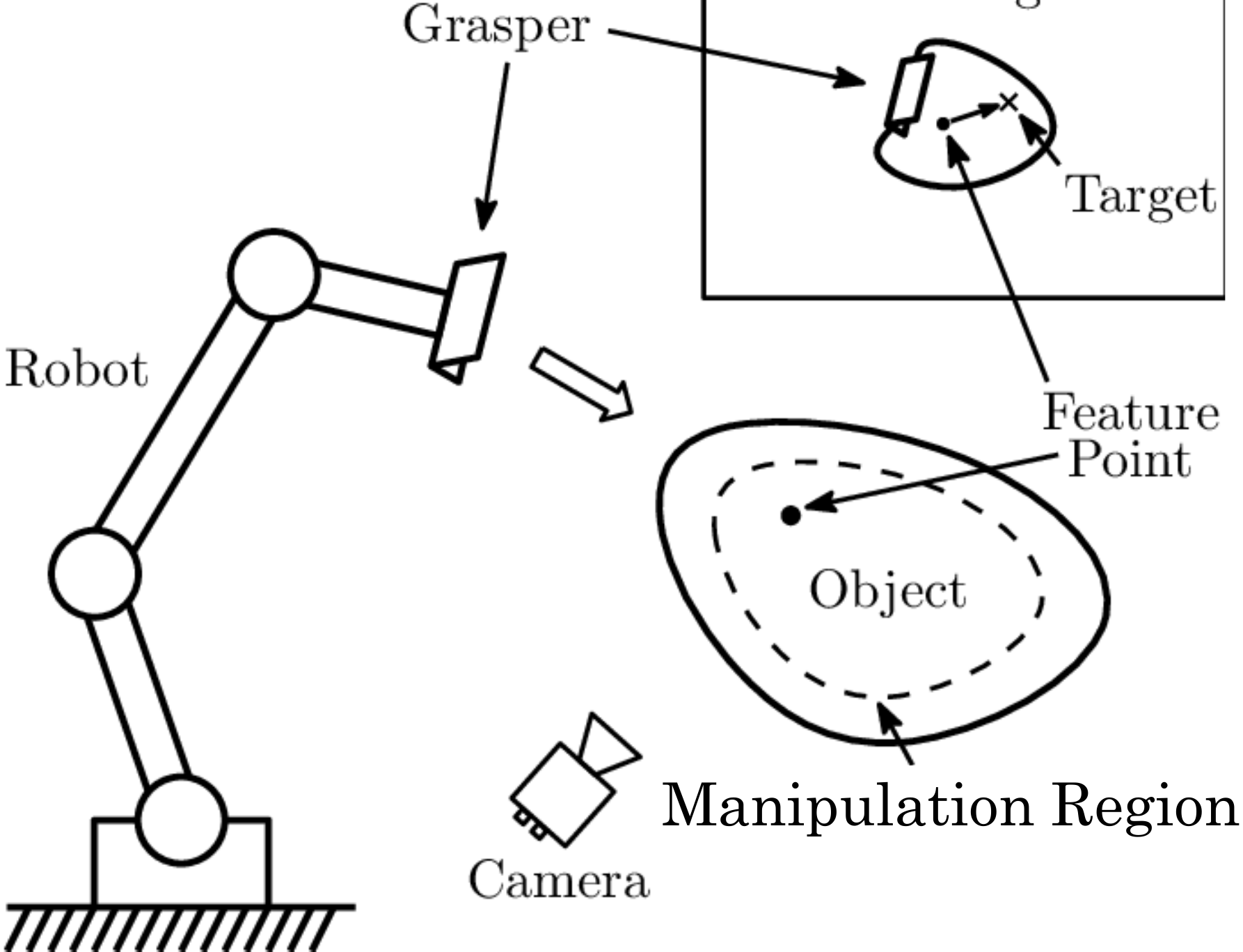}
    \caption{Manipulation region of an object}
    \label{fig:manipulation_region}
\end{figure}
 
To implement various relations among objects PyBullet allows to define \emph{manipulation regions} around objects. These regions are virtual regions inside which a robot can manipulate that object. A visual description of this region is shown in Figure~\ref{fig:manipulation_region}.

We now define two object types.
\begin{compactenum}
    \item \textbf{Standalone object:} Such an object is defined as a collection of stereo-lithographic tetrahedrons with visual and collision properties. Each such object has a single manipulation region, denoted as $MR(object)$, which is the region around the object where it is considered to be in ``vicinity'' to the object and hence can be manipulated by the robot. Examples: cubes, fruits, milk, etc.
    \item \textbf{Containing object:} Such an object is a standalone object with an additional containment region, denoted as $CR(object)$, which is a proper subset of the manipulation region where if another object’s center lies is called to be “contained inside” this object. Examples: box, cupboard and fridge.
\end{compactenum}

Every object has a center defined arithmetic mean of the centers of the tetrahedrons, denoted by $C(object)$. We now define the two relation constraints of which others are different versions of combinations of:

\begin{compactenum}
    \item \textbf{Inside:} An object $A$ is said to be contained inside a containing object $B$ if the $C(A) \in CR(B)$.
    \item \textbf{On Top:} An object $A$ is said to be on top on another object $B$ if $C(A)_z > C(B)_z$ and $MR(A) \cap MR(B) \neq \phi$, where $C(O)_z$ denotes the z component of the center of object $O$.
\end{compactenum}